\documentclass{article}

\usepackage[english]{babel}

\usepackage[letterpaper,top=2cm,bottom=2cm,left=3cm,right=3cm,marginparwidth=1.75cm]{geometry}
\usepackage{siunitx}
\usepackage{subcaption}

\usepackage{amsmath}
\usepackage{graphicx}
\usepackage[colorlinks=true, allcolors=blue]{hyperref}
\usepackage{placeins}
\usepackage[style=apa, backend=biber, url=false]{biblatex}
\usepackage{appendix}
\usepackage{pdflscape}
\usepackage{cleveref}
\usepackage{float}
\usepackage{xurl}
\usepackage{hyperref}

\usepackage{titlesec}

\title{Tropospheric temperature and humidity profile retrieval from Meteosat Flexible Combined Imager based on deep learning} %in Europe All-sky MTG-FCI 

\author{
Alejandro Salgueiro$^{1,2}$ \\
Johannes Rausch$^{3}$ \\
Julie Thérèse Villinger$^{3}$ \\
Angela Meyer$^{1,2}$ \\
\small$^{1}$ Department of Geoscience and Remote Sensing, Delft University of Technology, 2628 CN Delft, The Netherlands\\
\small$^{2}$ School of Engineering and Computer Science, Bern University of Applied Sciences, 2502 Biel/Bienne, Switzerland\\
\small$^{3}$ Meteomatics AG
}

\begin{document}
\maketitle

\begin{center}
\textit{Submitted for publication in Journal of Geophysical Research: Atmospheres.}
\end{center}

\section*{Key points}

\begin{enumerate}
\item A spatially aware U-Net retrieves forecast-independent all-sky tropospheric temperature and humidity profiles from MTG-FCI.

\item Validated against radiosondes, temperature STDs are below 2.0\,K and relative humidity within 20\,\% for all-sky retrievals in Europe.

\item The data-driven approach shows added value from short-wave infrared and near-infrared bands to constrain retrievals.

\end{enumerate}

\newpage
\begin{abstract}

The Meteosat Third Generation (MTG) Flexible Combined Imager (FCI) offers new opportunities for tropospheric temperature and humidity profiling, at higher spatio-temporal resolutions and expanded spectral coverage relative to its predecessor. Vertically resolved retrievals from broadband imagers are inherently challenging, and operational retrieval algorithms typically rely on numerical weather prediction (NWP) background fields to compensate for limited infrared spectral resolution, reducing the retrievals' independence. We develop a spatially aware deep learning framework to retrieve all-sky tropospheric temperature and humidity profiles from FCI, without forecast profiles as input. A Residual U-Net that exploits spatial context across all 16 FCI channels was trained on 14 months of collocated FCI observations and CERRA reanalysis targets over Europe. Validated against independent radiosondes, retrieved temperatures show biases below 0.4 K and standard deviations of 1.5–1.9 K. Retrieved relative humidity standard deviations range from 12–20\,\%, compared to 9–19\,\% for CERRA. Performance degrades modestly under clouds, with standard deviation increases below 0.4 K and 3\,\% RH beneath cloud tops despite limited direct radiative information. Ablation experiments show that spatial context improves retrievals, with the largest gains below cloud tops. Feature sensitivity analysis indicates broad consistency with FCI bands' established radiative transfer characteristics. Visible and near-infrared channels contribute despite not being commonly used in physics-based profile inversions. These results demonstrate that spatially aware deep learning models can extract statistically reliable tropospheric profiles from geostationary imager observations, independent of NWP forecast fields, enabling more rapid autonomous monitoring of the atmosphere.

\end{abstract}

\section*{Plain language summary}
Weather forecasting and climate monitoring depend on knowing how temperature and moisture vary through the atmosphere at different heights, information that is difficult to measure continuously over large areas. Weather satellites observe the Earth constantly, but for broad-spectrum imagers like the one studied here, converting satellite measurements into vertical atmospheric profiles has traditionally required inputs from weather forecast models to compensate for limited spectral detail. This study uses a deep learning model, an artificial intelligence trained on patterns in large datasets, to retrieve these profiles directly from Europe's newest geostationary satellite, Meteosat Third Generation's Flexible Combined Imager, without relying on forecast model output. The model exploits all 16 of the satellite's spectral channels, including near-infrared wavelengths less commonly used for this purpose, and draws on spatial patterns across the imagery. Trained on 14 months of satellite observations matched to atmospheric reanalysis data, and validated against independent radiosonde measurements, the method achieves temperature errors of around 1.5–1.9\,°C and relative humidity errors within 20\,\% across clear-sky and cloudy conditions. Performance degrades only modestly under cloud cover. These results show that spatially aware AI models can reliably extract atmospheric profiles from imager satellites, independent of forecast models.

\section{Introduction}

%Weather influences many aspects of modern society, from energy production and disaster risk management to agriculture and transportation. Accurate atmospheric monitoring therefore matters not only from a scientific perspective \parencite{desouzamachado_geophysical_2025} but also economically and socially. 
Satellite instruments provide global, frequent, and consistent atmospheric observations that in-situ networks cannot match, feeding numerical weather prediction (NWP) systems through two complementary pathways. Radiances can be assimilated directly, which avoids potential vertical resolution biases that may be introduced by profile retrievals. Alternatively, geophysical profile inversions can be assimilated or used directly for a range of nowcasting applications, such as instability monitoring or 3D atmospheric motion vectors estimation. In practice, however, direct assimilation schemes ingest only a small fraction of available channels, as the computational cost of running radiative transfer models at scale inside NWP systems forces operational centers to sub-select them. Offline retrieval algorithms can exploit the full spectral measurement vector and compress it into a compact form, recovering vertical information content reduced by channel sub-selection \parencite{migliorini_equivalence_2012} and motivating their continued development alongside direct assimilation.

Traditionally, profile retrievals have relied on physics-based inversion techniques, solving the radiative transfer equation and iteratively adjusting atmospheric state variables to minimise differences between observed and simulated radiances, typically within an optimal estimation framework constrained by NWP prior information \parencite{rodgers_inverse_2000}. Applying such methods to broadband imagers, however, presents significant physical and mathematical challenges. Unlike hyperspectral sounders with thousands of narrow spectral channels, imagers possess only a limited number of infrared bands with much wider spectral resolution. Their broad weighting functions average emitted radiation over thick atmospheric layers rather than discrete altitudes, severely limiting vertical resolving power. The resulting high interdependence between observations makes detailed profile retrieval a severely underdetermined and ill-posed problem, forcing retrievals to rely on NWP prior constraints to stabilise the solution, and consequently reducing the independence and added value of the retrieved profiles. Despite these limitations, operational profile retrievals from imagers do exist. NOAA's GOES-R atmospheric profiles product, derived from the Advanced Baseline Imager (ABI) \parencite{schmit_goes-r_2008}, uses a regression, combining measured radiances with NWP forecast profiles to guide an optimal estimation inversion. Imager-derived profiles such as those from ABI or the Flexible Combined Imager (FCI) \parencite{eumetsat_mtg-fci_2013} are primarily used to generate downstream products including integrated water vapour, instability indices, and atmospheric motion vectors. Nevertheless, \textcite{schmit_legacy_2019} demonstrated that ABI's NWP-constrained profiles can provide added value for short-range NWP forecasts, particularly for mid-to-upper tropospheric moisture. The growing data volumes of instruments such as FCI and the MTG Infrared Sounder (IRS) further strain traditional inversion pipelines, which already rely on fast but approximate radiative transfer models \parencite{saunders_update_2018} and channel sub-selection to meet latency constraints \parencite{august_iasi_2012}. More fundamentally, fast radiative transfer-based retrievals often make limited use of the full spectral information available. Visible, near-infrared, and shortwave infrared channels are often under-used due to the complexity of simulating solar scattering and surface emissivity effects \parencite{gao_water_2003}, and optimal estimation retrievals are often restricted to clear-sky conditions \parencite{li_satellite_2022}. These limitations create an opportunity for alternative approaches.

Data-driven methods have already demonstrated their ability to address several of these limitations simultaneously. A prominent example is the Piecewise Linear Regression (PWLR) method \parencite{eumetsat_validation_2021} used operationally by EUMETSAT for all-sky profile retrievals from IASI, which has delivered forecast-independent estimates complementing physics-based inversions for more than a decade. In practice, even traditional operational retrieval systems have incorporated data-driven components to provide initial state estimates, reducing NWP dependence and stabilising the inversion \parencite{seemann_operational_2003, schmit_goes-r_2008}. More recently \textcite{haynes_exploring_2023} demonstrated that deep learning, specifically a U-Net architecture, can learn the biases between short-range forecasts and radiosonde observations, combining forecast, ABI radiances and surface analysis. These successes motivate extending such concepts further with modern deep learning (DL) architectures. DL models can learn highly nonlinear relationships across spectral channels and, crucially, exploit spatial context from satellite imagery. This potential is already hinted at by \textcite{haynes_exploring_2023} and PWLR \parencite{eumetsat_validation_2021}, where both models leverage spatial context from a 3x3 patch to improve the central profiles.

This study was motivated by our expectation that, for geostationary imagers, larger spatial context could offer additional benefits. We expect that it can help compensate for missing radiative information below clouds, capture location-dependent thermodynamic structures and regional climatology encoded in the training data, and account for the slanted viewing geometry at high satellite zenith angles. An additional motivation is the potential we see for DL retrievals to naturally incorporate visible and near-infrared channels that are less often used in radiative transfer-based schemes, and because once trained, inference is near-instantaneous, making such frameworks well-suited to the latency constraints of both operational processing and real-time atmospheric monitoring. The Meteosat Third Generation imager, FCI, provides sixteen spectral channels spanning visible to infrared wavelengths, at unprecedented spatial and temporal resolution, which offer interesting synergy with the future IRS combined retrievals. While FCI alone cannot match the vertical resolving power of a hyperspectral sounder, its combination of spatial detail, 10-minute refresh rate, and new spectral channels, including the 0.9\,$\mu$m band with potential sensitivity to near-surface humidity features less accessible to infrared sounders, is unique among operational geostationary imagers. Current operational FCI profiles, which are not publicly available and only used for derived products, rely on NWP background fields and provide limited independent information \parencite{eumetsat_mtg-fci_2013}. No operational retrieval yet exploits the full spectral and spatial content of FCI imagery using modern deep learning techniques.

These gaps motivate the present work, which develops and evaluates a fully data-driven, spatially aware deep learning framework for retrieving all-sky tropospheric temperature and humidity profiles from FCI observations in a forecast-independent manner. We consider the framework as forecast-independent, as no prior profiles are used to guide the retrieval at inference time. A feature importance analysis further investigates how individual spectral channels contribute to the retrieval across vertical levels. This study is driven by two research questions:

\begin{enumerate}
\item How accurately can a fully data-driven, spatially aware deep learning approach retrieve all-sky three-dimensional tropospheric temperature and humidity fields from FCI observations in a forecast-independent manner?

\item Are the learned channel sensitivities consistent with radiative transfer expectations, and do solar channels contribute to measurable daytime skill?

\end{enumerate}

\section{Data and pre-processing}

\begin{figure}[h!]
\centering
\includegraphics[width=1.0\linewidth]{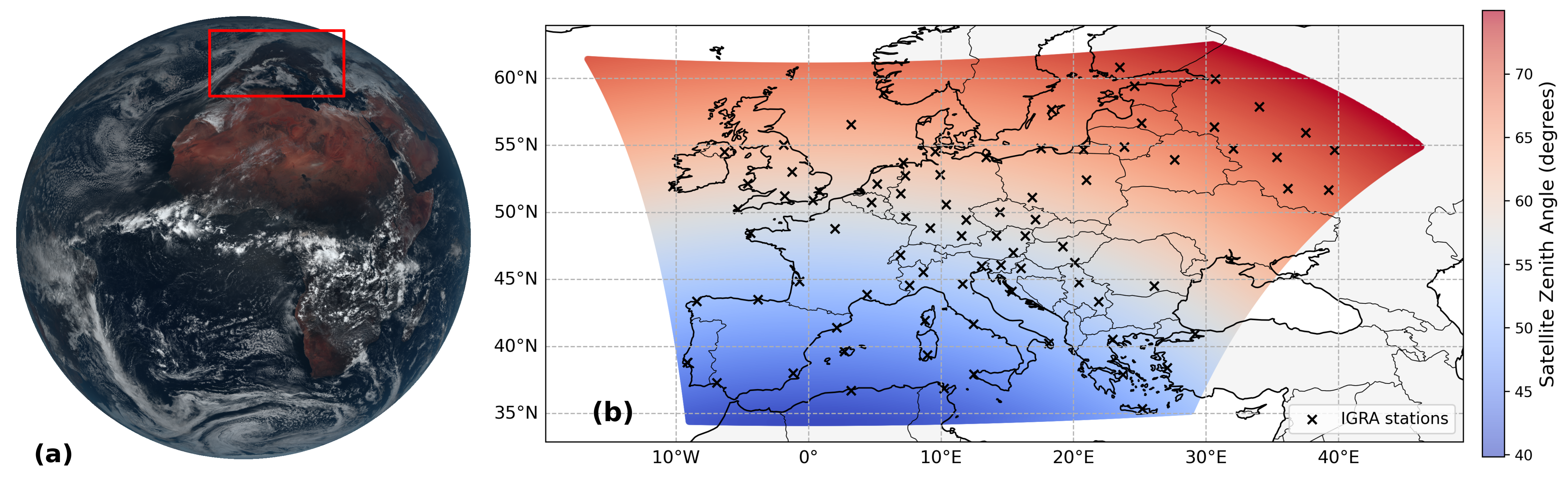}
\caption{
(a) True-colour RGB composite constructed from the FCI visible channels (0.4, 0.5, and 0.6\,$\mu$m). The red rectangle indicates the study domain. 
(b) Plate carree projection of the study domain map showing the locations of available IGRA radiosonde stations and satellite zenith angle. Values exceeding $75^\circ$ are masked to exclude large viewing geometries.
}
\label{fig:study_area}
\end{figure}

The primary model input consists of top-of-atmosphere observations measured by the FCI instrument aboard MTG, a geostationary satellite positioned at 0° longitude providing hemispherical coverage every 10 minutes across 16 spectral channels: 3 visible, 5 near-infrared, and 8 infrared, at nadir resolutions ranging from 0.5 to 2\,$\mathrm{km}$. Radiances were converted to reflectance for visible and near-infrared and brightness temperature for infrared observations, using Satpy FCI reader \parencite{martin_raspaud_pytrollsatpy_2026}. Unlike traditional radiative transfer-based retrieval pipelines \parencite{eumetsat_mtg-fci_2011}, which typically restrict the input to thermal infrared channels, we include the full set of 16 channels, allowing the model to exploit visible and near-infrared information on lower troposphere moisture, surface, cloud and aerosols characteristics. The study domain is restricted to Europe (Figure \ref{fig:study_area}), excluding pixels with satellite zenith angles above 75° where pixel spacing, atmospheric optical thickness and parallax effect for high altitude measurements become considerable (Supporting information Figures S1-S2). However, the presented retrieval approach is in principle also applicable to other domains. Since FCI scans from lower to higher latitudes every 10 minutes, and we use hourly observations, we therefore use the scan starting at hh:50. The scan is attributed to the full hour, introducing a temporal shift of 0.8-2.9 minutes depending on pixel location within the domain (Figure S1).

The retrieval targets are profiles of tropospheric air temperature and specific humidity from the CERRA regional reanalysis \parencite{ridal_cerra_2024} across the study domain. We use CERRA analysis fields, available at 5.5\,$\mathrm{km}$ horizontal resolution over Europe. Since reanalyses are available only every 3 hours, we also use short-forecast of CERRA's model at +1 and +2 hour to obtain hourly estimates, extracting air temperature and relative humidity for all available 15 pressure levels from 1000 to 300 hPa. We train the model to predict all pressure levels, even below surface pressure, but for evaluation we remove these measurements using CERRA surface pressure. Levels at pressures lower than 300 hPa are excluded and of less interest in this study as the absolute moisture content at those altitudes is very small. Rather than the 106 native model levels available in CERRA, this coarser vertical discretisation was chosen for computational tractability. Relative humidity in CERRA was converted to specific humidity using the \textcite{murphy_review_2005} parametrisations (Text S3).

The accuracy of the retrieval model is evaluated against independent radiosonde profiles from the National Oceanic and Atmospheric Administration (NOAA) Integrated Global Radiosonde Archive (IGRA) V2.2 database \parencite{durre_integrated_2016}, which aggregates measurements from multiple sources into a standardised format with integrated quality control. Only the highest-quality flagged measurements are used, log-linearly interpolated to the same 15 CERRA pressure levels. Profiles with a vertical gap exceeding 50 hPa between a radiosonde measurement and its nearest target level are discarded to avoid faulty radiosonde profiles with large data gaps. Radiosondes are used exclusively for evaluation and not for training. Spatial collocation assigns each radiosonde to its nearest FCI pixel, and temporal collocation accounts for the balloon ascent time. Since radiosondes take approximately 28 minutes on average to reach 300 hPa, as measured at Payerne station (Figure S4), the launch time is offset by 15 minutes before matching to the nearest hourly FCI observation, resulting in a maximum temporal shift of approximately 45 minutes. Operationally, FCI observations are available every 10 minutes, which would allow a much tighter temporal collocation, but for simplicity, we use hourly observations throughout this study. Horizontal drift during ascent for Payerne is also reported in the supporting information Figure S4. As a baseline reference, retrieval performance is also compared against a 30-year (1995-2025) ERA5 monthly climatology \parencite{copernicus_climate_change_service_era5_2019}, computed per hour of day for temperature, specific humidity, and relative humidity. We compare all-sky performances between retrievals, CERRA and ERA5 climatology against radiosondes.

Beyond FCI observations, we include a set of ancillary variables to provide the model with physical context that cannot be inferred from instantaneous observations alone. Surface elevation is provided by the Advanced Spaceborne Thermal Emission and Reflection Radiometer (ASTER) \parencite{nasa_aster_2019} and surface pressure by CERRA, corrected for topographic differences between datasets using a hypsometric adjustment (Text S3). The surface sun elevation angle is computed per observation using Astropy \parencite{the_astropy_collaboration_astropy_2022}. The time of day and day of year are encoded as sine-cosine pairs to preserve their cyclic nature. Finally, latitude, longitude, and satellite zenith angle are included as static spatial inputs, allowing the model to account for viewing geometry and regional climatological differences. While providing geographic coordinates risks increasing reliance on learned location-based climatology at the expense of radiance-driven skill, we accept this trade-off given that the model is not intended for use outside its training domain. All dynamic inputs are standardised to zero mean and unit variance, while static inputs are normalised to [0, 1] using min-max scaling. Visible and near-infrared bands' means and standard deviations were calculated only using measurements with a surface sun elevation above 0.0°, to avoid the distribution being dominated by near-zero nighttime noise values, which would bias the estimated mean and standard deviation away from the physically relevant daytime dynamic range of these channels.

FCI channels at 1\,$\mathrm{km}$ nadir resolution are spatially downsampled by a factor of 2 using non-overlapping 2×2 block averaging to match the 2\,$\mathrm{km}$ nadir resolution of the infrared channels. FCI data are distributed on a geostationary projection grid, where pixel footprints grow with distance from the sub-satellite point due to the increasing satellite zenith angle. Within the study domain, restricted to zenith angles below 75°, pixel areas range from 5.56 to 19.10\,$\mathrm{km}$² (Figure S1). All auxiliary gridded datasets (CERRA, ERA5, ASTER) are bilinearly interpolated to the FCI grid using the XESMF regridder \parencite{zhuang_pangeo-dataxesmf_2025}. The oblique viewing geometry also introduces a parallax displacement for elevated atmospheric features. A measurement from FCI at 10\,$\mathrm{km}$ altitude can appear shifted by up to 0.28° in latitude and 0.46° in longitude relative to its true surface position, corresponding to up to 4 pixels along FCI's y-axis and 3 pixels along the x-axis (Figure S3). Rather than applying an explicit parallax correction, we instead provide the model with spatial context by using large spatial contextual patches as input, allowing it to learn the correspondence between FCI observations and CERRA profiles across neighbouring pixels directly from the data. Our retrieval framework did not introduce any systematic errors for upper-tropospheric temperature and humidity at high satellite zenith angles (Figure \ref{fig:rmse_map}), which would indicate slanted, parallax-displaced retrievals, so we expect our model has implicitly learned to translate high satellite zenith angle radiances to coherent vertical structures similar to the reanalysis.

Clouds attenuate and block upwelling infrared radiation, preventing traditional radiative transfer retrievals from estimating the atmosphere below cloud tops. Rather than restricting the retrieval to clear-sky conditions, we train a single all-sky model to reconstruct full atmospheric profiles under both clear and cloudy conditions. We rely on measurements above and around clouds, partially clouded scenes, and the thermodynamic structures learned from CERRA to approximate temperature and humidity below cloud layers. Retrieval performance is therefore assessed separately for above- and below-cloud conditions in the evaluation. Cloud-top heights are obtained from the FCI Optimal Cloud Analysis (OCA) product \parencite{eumetsat_mtg-fci_2016}. For observations prior to 22 January 2025, when FCI OCA was not yet available, cloud-top heights from the Spinning Enhanced Visible and InfraRed Imager (SEVIRI) onboard the Meteosat Second Generation satellite are used instead. Cloud-top heights are corrected for parallax displacement, and the resulting spatial gaps are filled using a sliding 3×3 minimum kernel. This approach preserves spatial continuity while favoring the representation of lower cloud layers in the resulting field. The procedure is applied solely for evaluation purposes, specifically for distinguishing between above- and below-cloud retrievals, and introduces a conservative bias that may slightly under-represent the occurrence of high cloud-top profiles. Given that this correction affects only a small fraction of pixels, its impact on the overall analysis is considered limited.

Because the collocated dataset of 14 months is relatively short, the design of the training, validation, and test splits is critical. A simple chronological split would result in the training and test data originating from different seasons, thereby creating a potential domain mismatch between the training and test set. To address this, we use a rolling split strategy (Figure S5), following a repeating sequence of 21 days for training, 3.5 days for validation, 7 days for testing, and an additional 3.5 days for validation. This block structure limits the influence of atmospheric persistence and short-term synoptic autocorrelation between training and test periods, while successfully incorporating all seasons. The validation periods are placed closer to the training data because they are strictly used for model selection, such as hyperparameter tuning and early stopping, and do not directly affect the reported results. This main sequence yields a training set covering 252 days ($\sim$6048 scenes) and seasonally diverse test and validation sets of each $\sim$81 days ($\sim$1944 scenes), containing, for the test set, 11,840 valid collocated radiosonde profiles for performance evaluation.

To further assess generalization under minimal temporal autocorrelation and check for potential overfitting to long-timescale atmospheric structures, such as persistent large-scale circulation regimes, a secondary independent test set spanning from the 1st of December 2025 to the 28th of February 2026 is constructed, separated from the training period by a 7-day buffer of unused data. For this second test period, CERRA targets are unavailable due to a three-month dissemination delay; our retrieval framework instead uses ERA5-Land \parencite{copernicus_climate_change_service_era5-land_2019} over land and ERA5 \parencite{copernicus_climate_change_service_era5_2018} over ocean for surface pressure inputs, and is evaluated strictly against 12,419 independent radiosonde profiles. 

\section{Methods}

\subsection{Retrieval model}

The retrieval is formulated as a supervised image-to-image regression problem. The model receives as input a 128×128 pixel patch with 26 channels, 16 FCI spectral bands and 10 ancillary variables stacked along the channel dimension, and produces two output tensors of the same spatial size with 15 channels each, corresponding to temperature and specific humidity profiles at 15 pressure levels. This patch-based formulation is central to the approach. By operating on spatial neighbourhoods, the model receives spectral information to learn coherent three-dimensional temperature and humidity profiles, as well as cloud structures across the scene. At satellite zenith angles characteristic of Europe, the radiance observed at a given FCI pixel originates from an atmospheric column horizontally displaced from the pixel's surface footprint (Section 2). Training a pixel-wise model would ideally require reconstructing the slanted atmospheric column from CERRA's vertically-stacked profiles through horizontal interpolation across multiple grid points at each pressure level. To quantify the added value of spatial context, we nevertheless train a simplified pixel-wise baseline, without slant-column reconstruction, introduced below.

\begin{figure}[h!]
\centering
\includegraphics[width=1.0\linewidth]{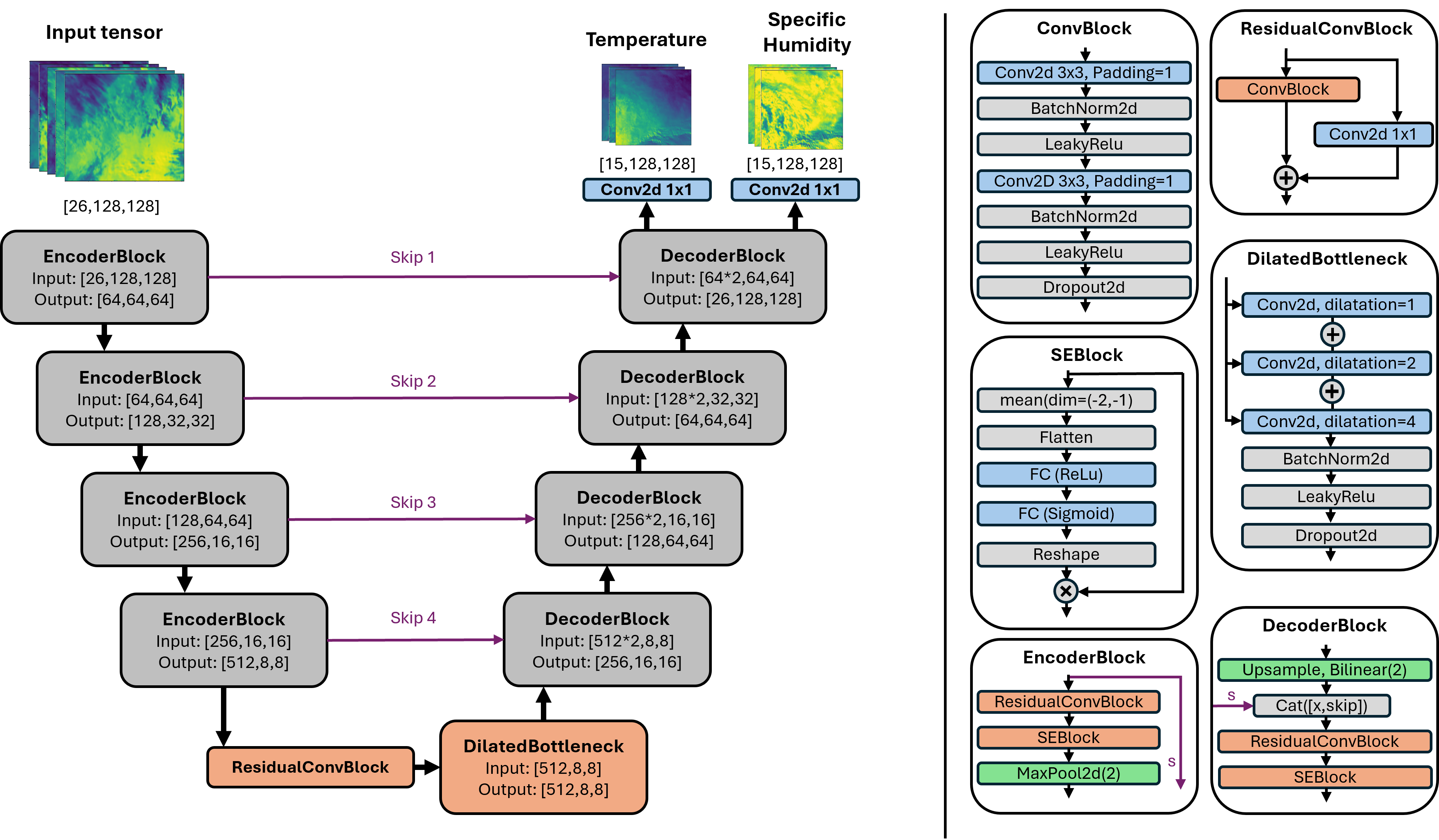}
\caption{\label{fig:unet} Residual U-Net with squeeze and excitation and dilated bottleneck blocks. In blue are the components with trainable parameters, in green the downsample and upsample components, in orange the main blocks and the rest of the components in gray.}
\end{figure}

The architecture of our main retrieval model, shown in Figure \ref{fig:unet}, is based on Residual U-Net (ResU-Net) \parencite{diakogiannis_resunet-_2020}, augmented with squeeze-and-excitation (SE) channel attention \parencite{hu_squeeze-and-excitation_2017} and a dilated multi-scale bottleneck inspired by DeepLab \parencite{chen_deeplab_2018}, which have both ben shown to improve validation loss and convergence speed compared to a simple ResU-Net. The encoder consists of 4 blocks, each combining a residual convolutional block with an SE attention mechanism \parencite{hu_squeeze-and-excitation_2017}, followed by $2 \times 2$ max pooling to downsample. Each residual convolutional block applies two successive $3 \times 3$ convolutions with padding 1, each followed by batch normalisation and a leaky ReLU activation. A learnable skip connection ($1 \times 1$ convolution when channel dimensions differ, identity otherwise) is added to the block output to facilitate gradient flow. The SE block performs channel-wise feature recalibration by globally average-pooling the spatial dimensions, passing the result through two fully connected layers each with a reduction factor of 16, and rescaling each channel by the resulting sigmoid-gated weights. The encoder progressively doubles the number of feature channels in each block except the first block which converts the 26 input channels into $C = 64$ base channel width. The encoder blocks have 64, 128, 256 and 512 channels which then go through the bottleneck. The bottleneck consists of a residual convolutional block followed by a dilated convolutional block. The dilated block applies three parallel $3 \times 3$ convolutions with dilation rates of 1, 2, and 4 (and corresponding paddings of 1, 2, and 4), summing their outputs before batch normalisation and leaky ReLU activation. This multi-scale receptive field encourages the model to integrate both local and broader spatial context at the deepest representation level. The decoder consists of 4 blocks that mirror the encoder. Each block bilinearly upsamples the feature map by a factor of 2, concatenates the corresponding encoder skip connection, and passes the result through a residual convolutional block followed by an SE attention block. Two separate $1 \times 1$ convolutional output heads finally transform the feature maps into two tensors of size $128 \times 128 \times 15$, corresponding to the predicted temperature and specific humidity profiles at 15 pressure levels. In total, the model amounts to 27.73 million parameters and 28.60 GFLOPS per forward pass. Dropout is applied at three levels: a rate of $p = 0.1$ in the encoder blocks, $p = 0.4$ in the bottleneck to regularise the most compressed representation, and $p = 0.3$ in the decoder blocks. 

To isolate the sources of retrieval skill, we additionally train two ablated baselines on the same CERRA targets, data splits, and training protocol. The first pixel-wise retrieval model, hereafter referred to as 1x1-Net, quantifies the contribution of spatial context: a counterpart of the U-Net in which all operations with spatial extent are removed. All $3 \times 3$ convolutions are replaced by $1 \times 1$ convolutions, and max pooling, upsampling, encoder--decoder skip connections, and SE attention are omitted. The resulting model corresponds to a deep residual MLP applied independently at each pixel, consisting of an input projection followed by 4 residual blocks of width 512, and otherwise retains the residual structure, batch normalisation, leaky ReLU activations, dropout ($p = 0.3$), and the two $1 \times 1$ output heads of the U-Net (2.5 million parameters). It is trained without the slant-column reconstruction discussed above; each output pixel thus depends only on the spectral channels of its own input pixel and cannot compensate for parallax displacement. The performance gap to the U-Net therefore reflects both the value of neighbouring spectral information and the geometric observation-target mismatch. The second baseline, hereafter ancillary-only, quantifies the skill attainable without any satellite observations: it is architecturally identical to the U-Net but receives only the 10 ancillary inputs, so its performance reflects spatio-temporal climatology and surface-pressure information alone, and the gap to the full model isolates the contribution of the FCI radiances. Both baselines converged more quickly than the U-Net; we use the 1x1-Net parameters at epoch 96 (of 116 trained) and the ancillary-only parameters at epoch 26 (of 46 trained), as the validation loss did not improve for a patience of 20 consecutive epochs.

The models are trained to minimise a Mean Squared Error (MSE) loss between predicted and target CERRA profiles, computed on standardised temperature and specific humidity values to equalise the contribution of both variables to backpropagation regardless of their differing physical scales. The use of different loss functions, notably losses promoting sharpness conservation, is left for future work and discussed in the section \ref{sect:spatial_results}. At each training step, 32 patches of 128×128 pixels are randomly sampled from the full spatial domain, with loss contributions from pixels above 75° satellite zenith angle masked out. We use an AdamW optimiser with a weight decay of $\lambda = 10^{-3}$ to mitigate overfitting, combined with a cosine annealing scheduler that decays the learning rate from $5 \times 10^{-5}$ to $5 \times 10^{-7}$ over a cycle of 100 epochs. The values of the hyperparameters were chosen to balance validation loss, convergence speed and overfitting. Training was halted at 146 epochs and weights taken from epoch 126, as the validation loss did not improve for a patience of 20 consecutive epochs.

\subsection{Evaluation}

For the U-Net and ancillary only model, full-domain retrieval fields are reconstructed by applying the trained model on overlapping 128×128 patches using a sliding window with a stride of 8 pixels (Figure S6a). Overlapping predictions are merged using Gaussian weighting that emphasises the central region of each patch (Figure S6b), which prevents artifacts near patch boundaries and modestly improved accuracy against radiosondes compared to a 128 stride window with no overlap. The 1x1-Net requires no aggregation, as the pixel-wise approach uses no spatial context and creates no boundary effects.

Retrieval performance is assessed for temperature, specific humidity, and the relative humidity derived from the two retrieved profiles, with all relative humidity values constrained to the range of 0-100\,\% to avoid supersaturation artifacts represented in radiosondes but not in CERRA \parencite{copernicus_climate_change_service_cerra_2022}. The primary metrics are bias (retrieval minus radiosondes) and standard deviation of the residuals (STD; Text S4), evaluated against independent radiosonde observations from IGRA. STD is preferred over the root mean squared error (RMSE; Text S4) as the headline metric because it isolates retrieval variability from systematic offsets, represented by the bias, that may also originate in the radiosonde or reanalysis, in addition to the retrieval framework itself, as discussed in Section 4.1.1. Performance is also classified by regions above and below cloud top, using FCI OCA, by solar illumination (sun elevation below 0° versus above 0°), and meteorological season (DJF, MAM, JJA, SON). Results from the second, temporally independent test set that covers DJF 2025/26 are compared against DJF 2024/25 performance from the main test set.

Radiosonde-based metrics quantify what \textcite{bonavita_forecast_2026} term statistical reliability, i.e. the ability to minimise expected error. However, such metrics can be inflated by overly smooth outputs that suppress gradients and reduce the ability to distinguish coherent meteorological structures, a property referred to as statistical resolution but that we broadly here term as sharpness. To assess sharpness, we complement point-based evaluation with a spatial analysis against CERRA over the full test domain. This includes visual inspection of randomly sampled two-dimensional fields and vertical cross-sections, a power spectrum analysis following \textcite{durran_practical_2017} to quantify scale-consistent spectral energy density and compare retrieval variability against CERRA across spatial scales (Text S5, Figure S7), and spatially distributed RMSE maps to identify systematic error patterns associated with viewing geometry, surface type, and geographic location over Europe. Since evaluation is performed against CERRA rather than radiosondes, systematic biases are shared between prediction and reference, and we therefore use RMSE.

\subsection{Feature importance analysis}
To evaluate how the model uses its input channels, we conduct a feature sensitivity analysis based on cross-sample channel substitution. Individual input channels are replaced by the corresponding channel from a randomly selected 128×128 patch drawn from a different time and location in the test set, preserving realistic marginal distributions while disrupting the physical consistency between channels for a given scene. The retrieval is recomputed for each perturbed input, and the resulting increase in RMSE relative to the unperturbed baseline, computed per variable and per pressure level against CERRA temperature and relative humidity profiles, serves as a measure of the model's sensitivity to perturbations in the input channels. The analysis is performed using 32 randomly located 128×128 patches per test-set timestep rather than the full reconstructed domain to maximise spatial diversity between samples. The resulting sensitivity estimates offer insight into how the retrieval draws on different spectral and auxiliary inputs across the troposphere, and allow a qualitative comparison with established radiative transfer theory of the FCI spectral channels.

This approach has two important limitations worth acknowledging. First, when input channels are correlated, the model can compensate for a substituted channel using correlated alternatives, leading to an underestimation of that channel's true contribution. Other attribution methods such as SHAP \parencite{lundberg_unified_2017} handle correlated inputs more robustly, but at considerably greater computational cost and with more complex interpretability for three dimensional outputs with different variability. We therefore retain the simpler substitution-based approach here. Second, neither method captures the contribution of information learned implicitly during training, regional climatology, surface-atmosphere coupling, large-scale dynamical regimes, and vertical thermodynamic consistency encoded in CERRA profiles. This analysis therefore reflects sensitivity to explicit input channels, while the role of learned background structure remains only indirectly represented and is not explicitly measured, but instead inferred from the interpretation of the results.

\section{Results and discussion}

\subsection{Statistical reliability of the retrieved profiles}

Interpreting errors in reanalysis-trained retrievals requires care: differences against radiosondes may reflect retrieval limitations or properties inherited from CERRA. Throughout this section, we therefore check whether error patterns persist when the retrieval is evaluated directly against CERRA, attributing shared structures to the training source. Results for CERRA profiles validated against radiosondes over the test period are provided in the supporting information (Text S6). Finally, all results should be interpreted in light of FCI limited data: the available 14 months of collocated data result in a test set of 81 days sampled in alternating 7-day blocks, amounting to 11.5 weeks and 11,840 radiosonde profiles, which limits meteorological representativeness.

\subsubsection{All-sky retrieval accuracy} \label{sect:all_sky_accuracy}

\begin{figure}[h!]
\centering
\includegraphics[width=1.0\linewidth]{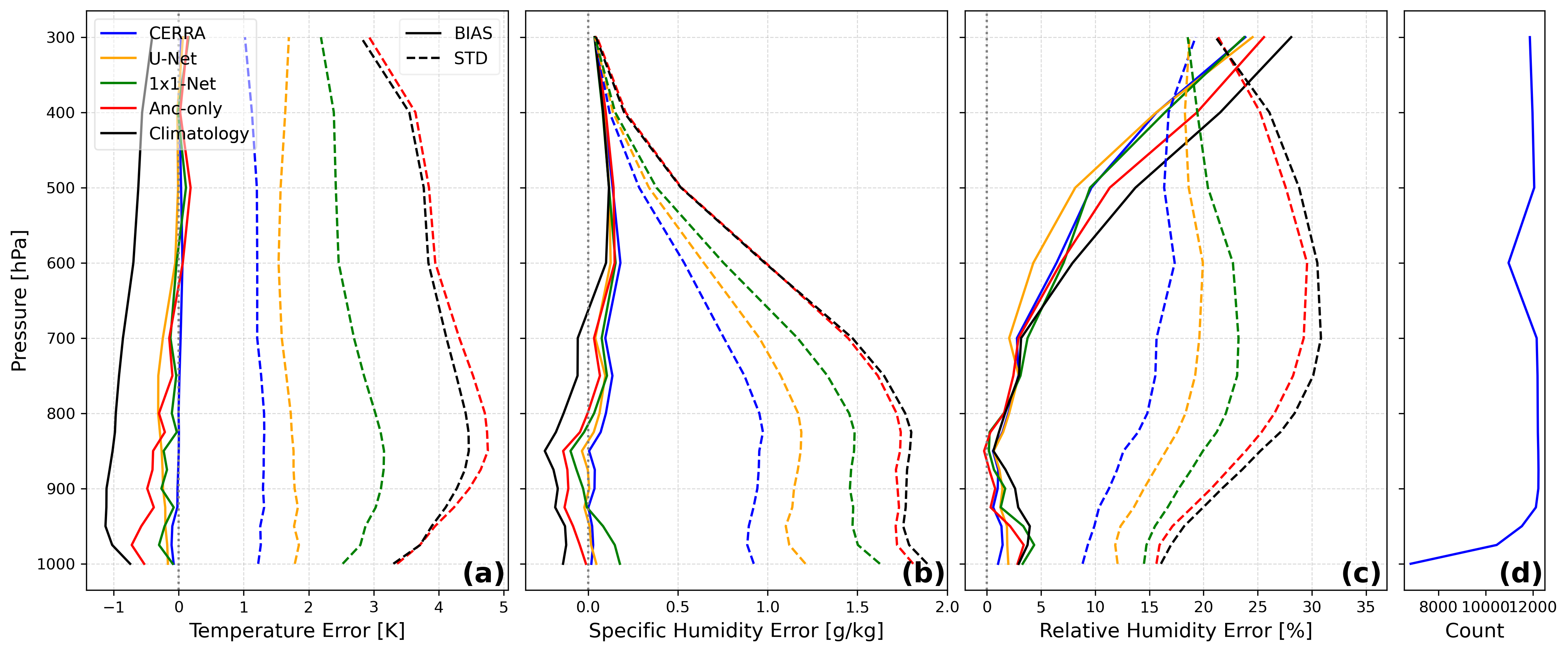}
\caption{\label{fig:nwp_ai_err} Test set bias and STD of profiles of (a) temperature, (b) specific humidity, and (c) relative humidity against radiosonde measurements in all sky conditions for CERRA, the U-Net, the 1x1-Net, the ancillary-only U-Net and ERA5 climatology. The subpanel (d) shows the total number of valid radiosonde measurements from all stations (Figure \ref{fig:study_area}b) used for each pressure level.}
\end{figure}

Figure \ref{fig:nwp_ai_err} summarises the vertical biases and STDs of air temperature, specific humidity, and relative humidity against radiosonde observations under all-sky conditions. The five products compared form a hierarchy of information content: the 30-year ERA5 climatology (a static mean using no observations), the ancillary-only model (adding location, time, and surface pressure), the 1x1-Net (adding single-pixel FCI radiances), the U-Net (adding spatial context), and CERRA (full data assimilation, and the training target). Differences between successive levels thus isolate the contribution of each component. The reduction in radiosonde measurements at 600\,hPa (Figure \ref{fig:nwp_ai_err}d) results from excluding interpolated radiosonde profiles where the nearest valid observation was more than 50\,hPa away. 

For temperature (Figure \ref{fig:nwp_ai_err}a), skill increases monotonically along this hierarchy. The climatology exhibits a pronounced cold bias of -1.2 to -0.5\,K. Both the climatology and the ancillary-only model perform similarly, with STDs reaching 4.5 to 4.7\,K. These results confirm that retrieval skill derives from the observed radiances rather than from learned spatio-temporal climatology. Adding single-pixel radiances reduces STDs to 2.2 to 3.2\,K (1x1-Net), and spatial context from the U-Net further to 1.5 to 1.9\,K, approaching CERRA's 1.0 to 1.4\,K. The U-Net retrieved biases remain below 0.5\,K, with a moderate cold bias at pressures higher than 600\,hPa. The slightly elevated variability at pressures higher than 700\,hPa is consistent with the known limitations of broadband infrared imagers in resolving more variable lower-tropospheric temperature and humidity through their broad vertical weighting functions \parencite{august_operational_2023}, but remains modest. Notably, the 0.7 to 1.3\,K added value of spatial context shows no marked amplification in the upper troposphere, where parallax displacement between observation and target is largest, suggesting the geometric mismatch contributes less to the 1x1-Net penalty than missing neighbourhood information, although radiosonde stations sample more moderate zenith angles (Figure \ref{fig:study_area}) and drift at higher altitudes (Figure S4), limiting the sensitivity of this test.

For specific humidity (Figure \ref{fig:nwp_ai_err}b), direct comparison across levels is complicated by the large vertical variation in absolute values, but the same ordering holds. Relative humidity (Figure \ref{fig:nwp_ai_err}c) shows it more clearly. The climatology and ancillary-only model again perform similarly, with STDs of 16 to 31\,\% with slightly lower STDs for the ancillary-only model (Figure \ref{fig:nwp_ai_err}b-c). The U-Net reaches STDs of 12 to 20\,\%, against 14.5 to 23.3\,\% for the 1x1-Net and 9 to 19\,\% for CERRA. The retrieved profiles from the U-Net and 1x1-Net perform particularly well in the upper troposphere, being competitive with CERRA at 300\,hPa where FCI's 6.3 and 7.3\,$\mu$m water vapour channels are most sensitive \parencite{schmetz_introduction_2002}.

All retrieved profiles, as well as CERRA and the ERA5 climatology, exhibit a notable positive relative humidity bias that grows at pressures lower than 700 hPa, reaching approximately 24\,\% at the highest levels. The similarity of the bias structure between these products, and with the climatology, warrants careful interpretation. While positive humidity biases have historically been attributed to radiosonde dry biases, modern European radiosonde models have mitigated the instrumental errors most responsible for this effect, such as radiation bias and time-lag errors \parencite{turner_dry_2003, ingleby_progress_2016}. Another contributing factor could be a structural overestimation of convective moisture transport in the upper troposphere within the reanalysis background \parencite{kunz_comparison_2014}, with reanalyses such as ERA40 exhibiting upper-tropospheric moist biases driven by errors in simulating the large-scale circulation \parencite{chung_model-simulated_2011}. Positive biases of comparable scale have been reported from other satellite retrievals against radiosondes, including the Advanced Baseline Imager against the same \parencite{schmit_legacy_2019}, and IASI against IGRA radiosonde data \parencite{eumetsat_validation_2021}. Correction pathways exist depending on the source: if the bias originates primarily from the reanalysis, fine-tuning directly against radiosondes offers a direct solution; if it is shared across datasets, bias correction using independent humidity measurements, as demonstrated by \textcite{wang_machine_2025}, provides an effective alternative. Given that these biases seem consistent between retrieval and reanalysis, we consider that the STD is the more informative metric for assessing humidity retrieval performance.

Overall, the retrieved profiles perform respectably across temperature and humidity under all-sky conditions, substantially outperforming both the climatological lower bound and the 1x1-Net while approaching CERRA accuracy, without relying on an NWP background state. The consistent moist bias structures between the retrieved profiles and CERRA point to shared origins rather than retrieval-specific errors, and represent a tractable target for future refinement. 

\subsubsection{Above vs. below cloud tops}

\begin{figure}[h!]
\centering
\includegraphics[width=1.0\linewidth]{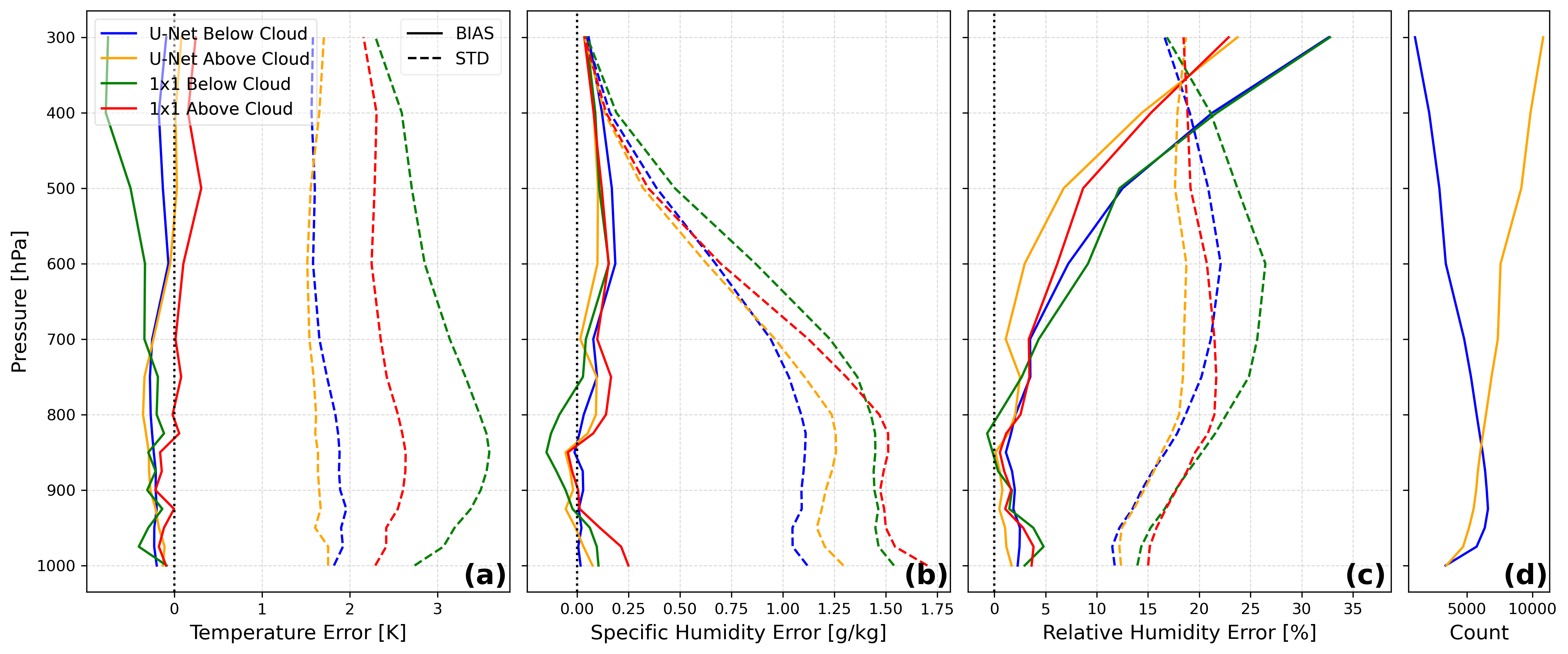}
\caption{\label{fig:ai_cloud_err} Test set bias and STD of retrieved profiles of (a) temperature, (b) specific humidity, and (c) relative humidity against radiosonde measurements for above and below clouds with the U-Net and 1x1-Net. Blue and green lines represent the errors below cloud while orange and red above the cloud top. 
The subpanel (d) shows the total number of valid radiosonde measurements used for each cloud condition.}
\end{figure}

Figure \ref{fig:ai_cloud_err} separates retrieval performance into above-cloud and below-cloud regimes, defined using the cloud-top pressure from the FCI OCA product \parencite{eumetsat_mtg-fci_2016}. The sample distribution in the rightmost panel (Figure \ref{fig:ai_cloud_err}d) reflects the natural prevalence of each regime: below-cloud samples dominate near the surface up to 850 hPa, while above-cloud samples become increasingly prevalent toward the upper troposphere. This sampling asymmetry should be kept in mind when interpreting level-by-level statistics. We evaluate the retrieval performance partitioned into above-cloud and below-cloud regions, rather than simply clear-sky and overcast conditions. Because dense clouds act as an opaque infrared boundary, traditional physical retrievals can still estimate temperature and humidity profiles down to the cloud-top level by explicitly redefining the lower boundary condition of the radiative transfer equation \parencite{zhou_physically_2007}. We expect our data-driven model to implicitly learn this same capability, accurately extracting above-cloud temperature and humidity structures. The model, using the climatology learned from CERRA, should then be able to use this information to reconstruct below-cloud profiles, where the spectral information is attenuated. Throughout this section, the 1x1-Net serves as a diagnostic: since it lacks any neighbourhood information, comparing the two models by regime separates the contribution of spatial context from constraints available to a pixel-wise retrieval.

For temperature (Figure \ref{fig:ai_cloud_err}a), biases remain below 0.5\,K in both above-cloud and below-cloud regimes. STDs remain within 1.5 to 2.0\,K across the profile, with above-cloud retrievals showing lower variability in the lower troposphere, consistent with the reduced radiative information below cloud tops. This variability difference narrows in the upper troposphere, with plausible explanations lying in two mechanisms. First, cloud tops rarely extend into the upper troposphere, reducing the vertical distance of the measured cloud characteristics (temperature, height, phase, optical thickness) and the higher altitude measurements, providing additional local context. Second, upper-tropospheric clouds are often more dispersed, consist predominantly of ice crystals, and are optically thinner in the thermal infrared than liquid water clouds. This could allow a fraction of upwelling radiances from below to reach the instrument \parencite{zhou_physically_2007}. The radiative transfer implications of cloud-affected satellite radiances represent an important direction for ongoing research \parencite{geer_allsky_2018,li_satellite_2022} which can also be explored through data-driven all-sky retrieval approaches. Overall, the retrieved profiles maintain consistent temperature accuracy in both regimes, suggesting that learned vertical consistency with retrieval above-cloud and spatial context may contribute to below-cloud accuracy beyond climatological smoothing. The 1x1-Net sharpens this picture: its above/below-cloud STD gap is substantially larger than the U-Net's, with below-cloud temperature STDs degrading by 0.7 to 1.7 K and above-cloud STDs within 0.4 to 1.0 K of the spatial model. The added value of spatial context is thus more important where direct radiative information is blocked, consistent with neighbouring clear and thin-cloud pixels supplying the constraint that a single attenuated pixel cannot. In the upper troposphere, 1x1-Net biases grow, becoming colder below cloud tops and warmer above them.

For specific humidity (Figure \ref{fig:ai_cloud_err}b), regime differences are more pronounced. Biases remain small in the lower troposphere at pressures higher than 700 hPa in both conditions, but below-cloud retrievals have somewhat higher biases at higher altitudes, a pattern shared with CERRA against radiosondes (Figure S8), pointing to some possible inheritance from the training source. In terms of variability, below-cloud retrievals are competitive with, and in the lower troposphere even outperform, above-cloud retrievals. One possible explanation is that, at pressures higher than 700\,hPa, the lower troposphere below clouds tends to be both well-mixed and moister, narrowing the range of likely humidity states, which could help constrain retrievals even without direct radiative information. Notably, this below-cloud advantage in the lower troposphere persists in the 1x1-Net, supporting a state-space rather than spatial-context origin: the narrowed range of likely boundary-layer humidity states constrains even a retrieval with no neighbourhood information. At pressures lower than 650 hPa, above-cloud retrievals show lower STDs than below-cloud. Cloud presence at these altitudes is often associated with convective updrafts and larger moisture variability between levels \parencite{wallace_atmospheric_2006}, which are difficult to retrieve without direct radiative sensitivity, while above-cloud retrievals benefit from FCI's water vapour channels being most sensitive in the mid-upper troposphere \parencite{schmetz_introduction_2002}. Relative humidity (Figure \ref{fig:ai_cloud_err}c) shows broadly consistent patterns with specific humidity. Biases grow at pressures lower than 700 hPa, reaching approximately 23\,\% for above-cloud and 33\,\% for below-cloud retrievals. Above-cloud STDs (12 to 19\,\%) are generally lower than below-cloud STDs (11 to 23\,\%) through the mid troposphere, with the gap narrowing at pressures lower than 600 hPa, possibly for the same reasons discussed for temperature: additional information from vertically nearby cloud top and reduced infrared attenuation from sparse and optically thinner clouds. The 1x1-Net again shows larger relative humidity STDs, degrading by 0.2 to 4.6\,\% below cloud and by up to 3.5\,\% above cloud, and similarly to temperature, humidity errors structure mirrors the U-Net's, with a slightly larger above/below-cloud gap.

These regime differences persist when the retrieval is evaluated directly against CERRA (Figure S9), indicating some above- versus below-cloud differences beyond inherited structure from the training source. In summary, above-cloud temperature retrievals are slightly better in the lower troposphere but otherwise comparable to below-cloud retrievals, while humidity shows a more pronounced regime dependence. The 1x1-Net comparison disentangles the contributing mechanisms: spatial context provides its largest gains below cloud tops, where it seems to compensate for blocked radiative information, whereas the below-cloud constraint on lower-tropospheric humidity reflects the narrower range of boundary-layer states and is available even to a pixel-wise model.

\subsubsection{Seasonal and diurnal variability}

Retrieval performance varies with meteorological season, as detailed in Figures S12–S14 of the supporting information. Due to the 14 months of data and the rolling split (Figure S5), the seasonal distribution is uneven: SON provides the most training samples (2297 hourly scenes), followed by MAM (1387), JJA (1231) and DJF (1133), possibly favouring SON results. Temperature STDs follow a winter-to-summer gradient consistent with the greater baroclinic variability and cloud cover of the European winter, while specific humidity STDs mirror the seasonal absolute moisture cycle. These patterns broadly align with seasonal behaviour reported for IASI over Europe \parencite{eumetsat_validation_2021} and persist when comparing the retrieval against CERRA (Figure S12), suggesting they reflect genuine atmospheric variability rather than inheritance of the training data.

Retrieval performance also varies with illumination (Figure S13). Daytime retrievals outperform nighttime across the profile, with temperature STDs lower by 0.5 to 1.2\,K, relative humidity STDs by 1 to 5\,\%, and similar relative reductions for specific humidity; biases remain below 0.5\,K for temperature in both conditions, with slightly elevated daytime wet biases for specific humidity.
With illumination defined by solar elevation, nighttime samples are weighted toward winter, so seasonal composition partly enters this comparison, but in opposite directions for the two variables. As discussed above, for temperature, winter's stronger baroclinic variability and cloud cover independently degrade retrievals, inflating the apparent daytime advantage. For specific humidity, winter's lower absolute moisture reduces STDs, so the daytime advantage persists despite, rather than because of, the seasonal weighting. Additionally, part of the remaining daytime advantage could be inherited from the training source: CERRA itself performs markedly better against radiosondes during daytime, with an even larger day-night gap than the retrieval (Figure S14). Contributing factors may include denser daytime observation usage in the assimilation, e.g. commercial aircraft measurements concentrated in daytime flight hours \parencite{ridal_cerra_2024}, and the difficulty of representing stable nocturnal boundary layers for reanalysis and retrieval alike. Consistently, when the retrieval is evaluated directly against CERRA, which removes target quality and largely controls for seasonal composition, since retrieval and reference share the same samples, the day-night gap largely disappears for specific humidity (Figure S15), while a residual gap persists for daytime temperature retrievals. These entangled comparisons cannot cleanly separate training-data quality, seasonal composition, and retrieval skill, but the most direct evidence for the added value of the visible and near-infrared channels remains the feature importance analysis in Section 4.3. \textcite{wagner_use_2024} reported only a slight daytime humidity improvement for IASI level 2 retrievals over Europe, attributed to warmer surfaces and stronger thermal contrast favourable to sounding; our substantially larger differences suggest that thermal contrast may contribute to the daytime advantage but is insufficient alone to explain it.

\subsubsection{Temporally independent generalisation}

\begin{figure}[tbhp]
\centering
\includegraphics[width=1.0\linewidth]{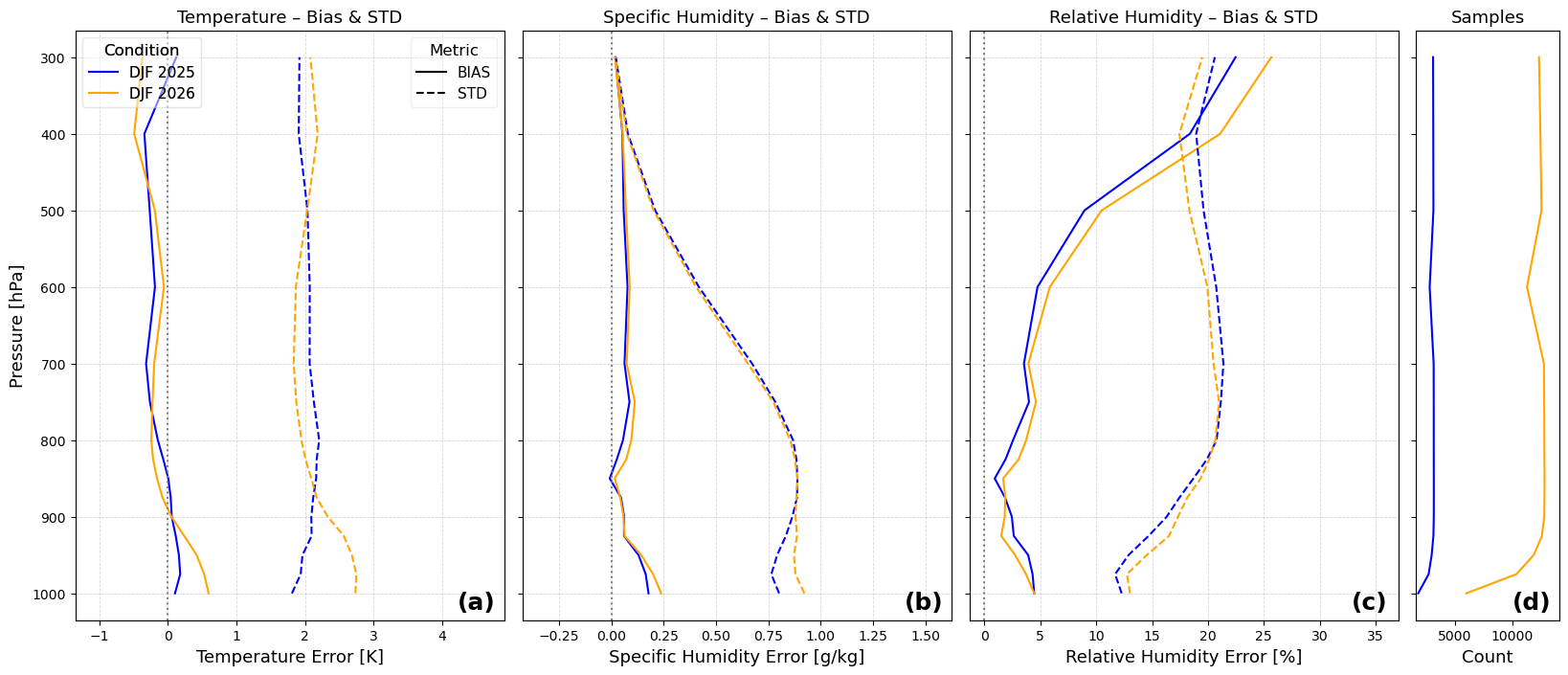}
\caption{\label{fig:ai_nwp_err_2} Bias and STDs of our retrieved profiles for the first and second test sets against radiosondes. Blue: main test set (DJF 2024/25), using the rolling-split test days (Figure S5). Orange: second test set (DJF 2025/26), using all hourly retrievals.}
\end{figure}

A natural concern for any data-driven retrieval trained on a limited record is whether the model learns the atmosphere or merely its training set. The alternating train-validation-test structure of the data split (Figure S5) enforces a minimum temporal separation of 3.5 days between training and test samples, reducing the likelihood that nearly identical synoptic conditions appear in both sets. As a stricter test, the retrieval is also evaluated on a second period (DJF 2025/26) lying entirely outside the training window
(Figure \ref{fig:ai_nwp_err_2}). This second test set yields a larger radiosonde sample, as it is not subject to the rolling split and all hourly retrievals are used; it also relies on ERA5 rather than CERRA surface pressure as input (Section 2), a known difference to keep in mind when interpreting the comparison.

Retrieval skill is well preserved across the mid-troposphere (400 to 900\,hPa), where biases and STDs remain closely comparable between the two periods. The most notable degradation occurs near the surface, at pressures higher than 900\,hPa, where temperature STDs increase by 0.2 to 0.9\,K and specific humidity STDs by 0.01 to 0.12\,g/kg, accompanied by warmer and moister biases. Smaller temperature degradations appear at 400-300\,hPa (STD increases up to 0.27\,K), while relative humidity generalises particularly well, with only marginally moister biases and STD differences of 0 to 3\,\% across the full column.

The vertical specificity of this degradation is itself informative: a retrieval that had overfit to specific synoptic events would be expected to degrade more uniformly across the column. Instead, the localised near-surface loss points to the meteorological
representativeness of the 14-month training record. The DJF 2025/26 period contains more cold and dry near-surface radiosonde observations, below $-10$\,\textdegree C and 1\,g/kg at 950 and 1000\,hPa (Figures S16-S17), conditions under-represented in training and possibly overestimated by the model; the ERA5 surface pressure input may contribute further. Extending the training
record as the FCI archive accumulates, exploring domain adaptation from historical SEVIRI observations as a pre-training source, or including ground station observations to constrain the boundary layer \parencite{huang_optimizing_2025} represent different paths toward improved generalisation.

\FloatBarrier

\subsection{Spatial structure and resolution of retrieved fields} \label{sect:spatial_results}
Until now, the evaluation has focused on the U-Net's ability to minimise profile variability as bias and STDs against radiosondes, a property defined as \textit{statistical reliability} \parencite{bonavita_forecast_2026}. Next, we characterise the model's capacity to detect and differentiate atmospheric structures, which we refer to as \textit{sharpness}.

\subsubsection{Retrieved field structure}
Figures \ref{fig:retrieval_2d_300}-\ref{fig:retrieval_2d_900} present retrieved temperature and specific humidity fields at 300 hPa and 900 hPa for a randomly selected test case, alongside the corresponding CERRA fields, FCI's 7.3\,$\mu$m water vapour channel brightness temperature, and the cloud top pressure (CTP) OCA product \parencite{eumetsat_mtg-fci_2016}. Large-scale structures are properly reproduced at both pressure levels, which is encouraging for 900 hPa humidity, where FCI's infrared water vapour channels have little sensitivity. 

The retrieved fields are nonetheless noticeably smoother than CERRA, exhibiting less pronounced gradients. We expect this can be explained by multiple reasons. First, cloud cover, visible through the CTP product, blocks radiative information from below cloud top entirely, producing the smoother gradients visible in the upper-right portion of Figures \ref{fig:retrieval_2d_900}a and \ref{fig:retrieval_2d_900}c where the scene is predominantly overcast (Figure \ref{fig:retrieval_2d_900}f). Furthermore, the broad vertical weighting functions of all bands further limit both vertical and horizontal resolution, as the model tends to spatially smooth retrievals where channel sensitivity is ambiguous. Additionally, while aggregating overlapping patches with Gaussian weights helps eliminate edge artifacts, it likely introduces some spatial smoothing as a side effect (Figure S6). Finally, a limitation common to MSE-optimised retrievals, is the systematic attenuation of sharp gradients and extreme values. MSE minimisation doubly penalises sharp, spatially offset features, once for missing the event at its true location and once for predicting it where none occurred, making attenuated averages safer than sharp gradients. The loss function choice shapes retrieval properties since it affects the trade-off between statistical reliability and sharpness. Adapting a loss function to the intended application is therefore an important next step in the development of spatially aware deep learning retrieval frameworks.

Finer structures emerge in zoomed-in fields (Figures \ref{fig:retrieval_2d_300}g-j and \ref{fig:retrieval_2d_900}g-j), seemingly inheriting smaller scale from FCI measurements. In Figures \ref{fig:retrieval_2d_300}g and \ref{fig:retrieval_2d_900}g for example, we can see small similarities in patterns between temperature and the colder brightness temperature (Figure \ref{fig:retrieval_2d_300}k) on the bottom right and near the higher cloud (Figure \ref{fig:retrieval_2d_300}l) borders. The smoother appearance of CERRA is likely affected by both the bilinear interpolation, and the numerical dissipation, spatial filters and assimilation damping of CERRA \parencite{ridal_cerra_2024}.

\begin{landscape}
    \begin{figure}[h!]
    \centering
    \includegraphics[width=1.0\linewidth]{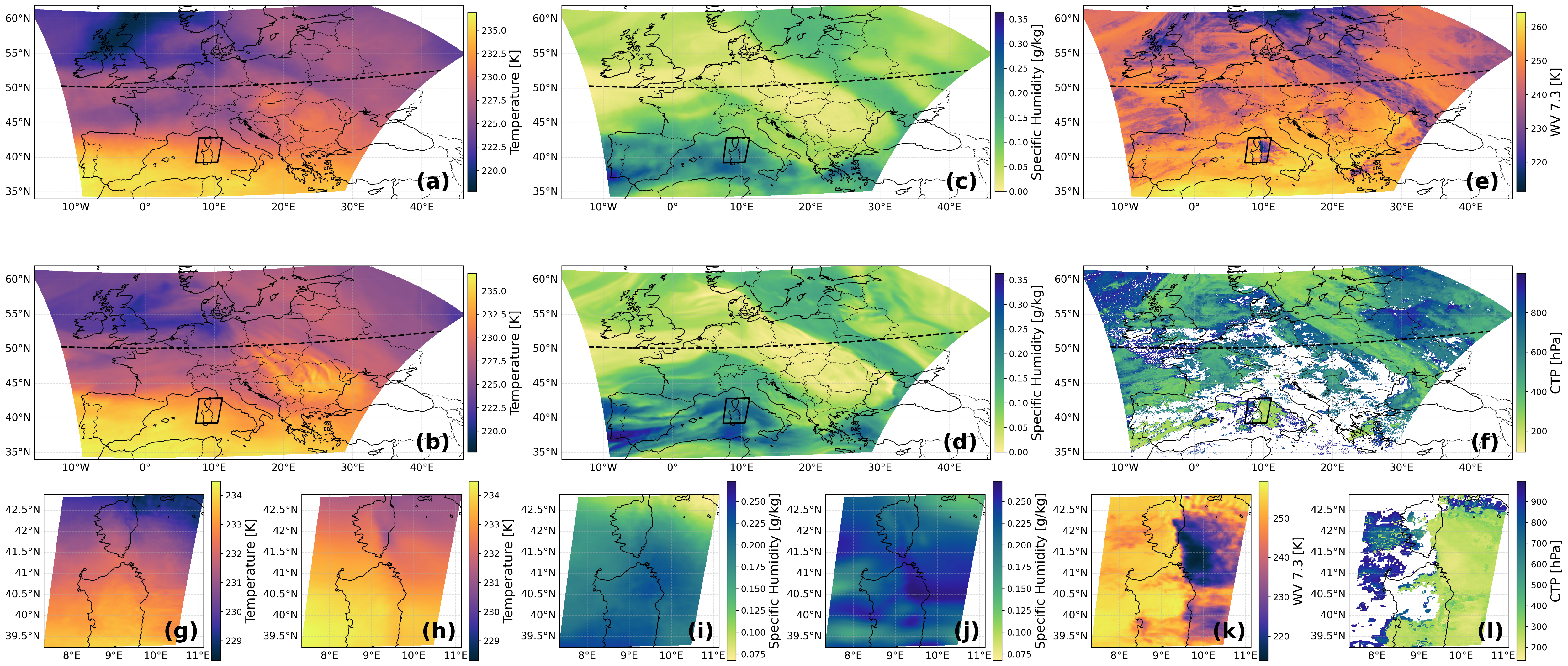}
    \caption{\label{fig:retrieval_2d_300} Spatial validation of retrieved atmospheric profiles against CERRA reanalysis data at 300 hPa on October 23, 2024 at 0600 UTC. Panels (a-f) illustrate the full spatial domain, and panels (g-l) display the target zoomed region marked by the solid black rectangle. Columns compare retrieved (a,g) and target (b,h) temperature, retrieved (c,i) and target (d,j) specific humidity, alongside contextual satellite inputs: FCI Water Vapor channel 7.3\,$\mu$m (e, k) and Cloud Top Pressure (f, l). The black dashed line indicates the cross-section track analyzed in Figure \ref{fig:cross_section}. 
    }
    \end{figure}

    \begin{figure}[h!]
    \centering
    \includegraphics[width=1.0\linewidth]{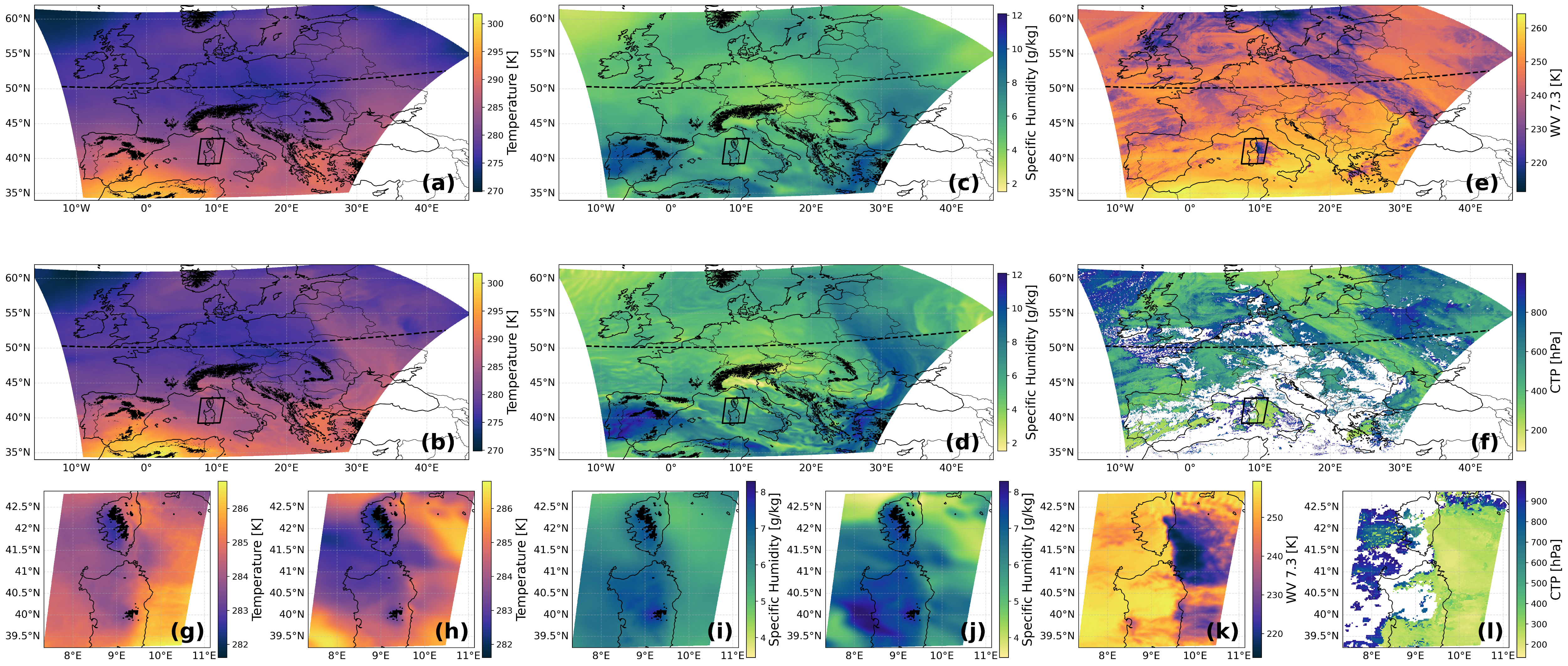}
    \caption{\label{fig:retrieval_2d_900} Same as Figure \ref{fig:retrieval_2d_300} for the 900 hPa pressure level. Areas where the measurements are below pressure levels are masked in black.
    }
    \end{figure}
\end{landscape}

\begin{figure}[h!]
\centering
\includegraphics[width=1.0\linewidth]{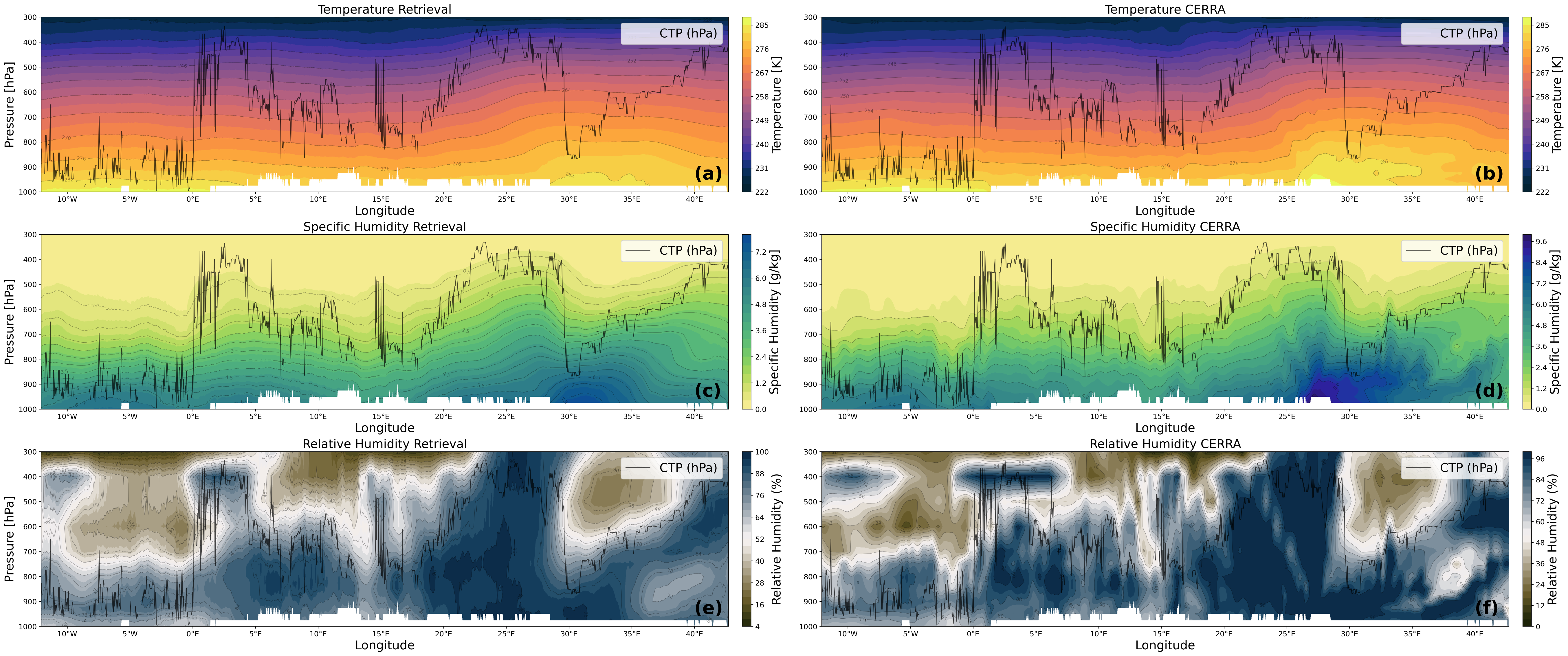}
\caption{\label{fig:cross_section} Vertical cross-section comparison between retrievals and CERRA for temperature (a-b), specific
humidity (c-d), and relative humidity (e-f) along a selected transect. The columns show retrievals (left)
and CERRA fields (right). Purple lines indicate the cloud top pressure along the transect.}
\end{figure}

Figure \ref{fig:cross_section} presents vertical cross-sections of retrieved temperature, specific humidity, and relative humidity along the dashed line shown in Figure \ref{fig:retrieval_2d_300}-\ref{fig:retrieval_2d_900}, compared against CERRA. The cross-sections confirm the picture emerging from the 2D fields while offering additional insight into vertical structure.
Large-scale patterns are well reproduced across all three variables, and the retrieved moisture content broadly follows the cloud top height along the section. In the relative humidity field shown in Figure \ref{fig:cross_section}e-f, the retrieval shows some awareness of two different vertical humidity layers between $\sim12^\circ-8^\circ W$ and $\sim0^\circ-5^\circ E$. This seems consistent with FCI's documented capability to distinguish two cloud layers when the upper one is optically thin \parencite{eumetsat_mtg-fci_2016}. By analyzing a large spatial context, the U-Net should also be able to identify spatially close humidity or cloud gradients, and structural transitions, drawing on its learned climatology to infer the vertical interaction and alignment of overlapping humidity layers. However, the warm and moist anomaly in the lower troposphere between 25°and 35°E illustrates the limits of the approach more directly. The warmer and moister conditions are partly captured, but with smoother gradients and reduced intensity. Cloud cover above this feature limits the available radiative signal, but even in clear-sky conditions the combined effect of broad weighting functions and MSE optimisation would prevent sharp, intense lower-tropospheric anomalies from being sharply resolved. The relative humidity field also exposes a specific gap. Despite broadly following cloud top structures, the retrieval rarely approaches saturation. This may be a consequence of optimising only temperature and specific humidity MSE during retrieval model training, smoothing important values and gradients. Accounting for relative humidity and near-saturation in the optimisation with additional weighting near cloud locations is a possible open direction for future research to improve the representation of humidity around clouds.

\subsubsection{Scale-dependent variability}
\begin{figure}[h!]
\centering
\includegraphics[width=1.0\linewidth]{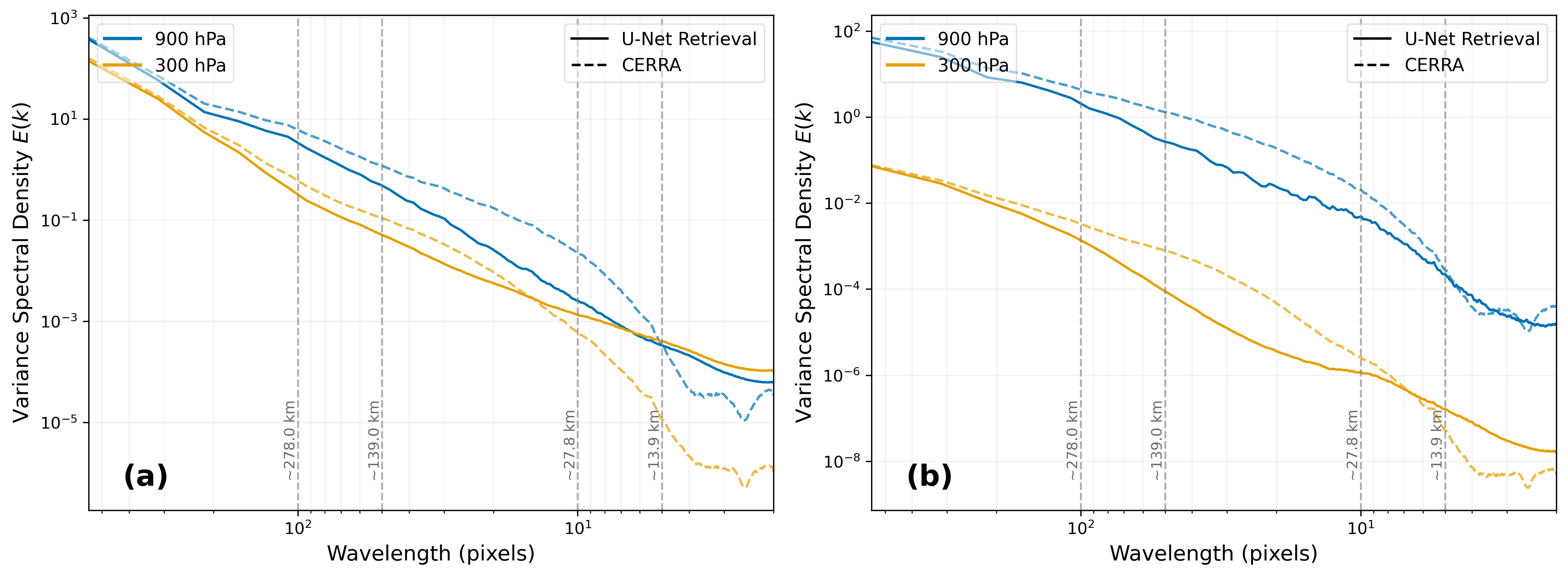}
\caption{\label{fig:power_spectrum_analysis} Radially averaged dimensional variance spectral density of (a) temperature and (b) specific humidity as a function of spatial wavelength (in pixels) for selected pressure levels. Solid lines represent the U-Net retrieval, and dashed lines correspond to the CERRA reference targets. Vertical dashed lines indicate equivalent physical distances based on the average grid spacing of 2.78\,km/pixel. Technical details of the spectral computation are provided in the supporting information Text S5.}
\end{figure}

To quantify spatial variability across scales, we analyse the power spectra of the retrieved and CERRA temperature and specific humidity fields. Because the analysis is performed on the satellite's native geostationary grid, pixel footprints expand at higher viewing angles (Figure S7), introducing some spectral smearing. However, since both the retrieval and CERRA are evaluated on the same non-linear grid, their relative comparison remains consistent. To simplify interpretation, we approximate FCI grid resolution using the square root of the mean pixel area across the cropped domain (Figure S7), giving an average resolution of 2.78\,$\mathrm{km}$ per pixel, but we keep the power spectrum Figure \ref{fig:power_spectrum_analysis} in pixels.

Across all variables and pressure levels, the power spectra share a common structure (Figure \ref{fig:power_spectrum_analysis}): CERRA carries increasingly more energy than the retrieval from the largest scales down to approximately $\sim10-20\,pixels$ ($\sim28-55\,\mathrm{km}$). This confirms quantitatively what was visually apparent in the 2D fields (Figures \ref{fig:retrieval_2d_300}-\ref{fig:retrieval_2d_900}): Synoptic-scale structures are broadly captured, but mesoscale gradients are smoothed as a consequence of the limited spectral information available to FCI and the attenuation of MSE optimisation. At around $\sim28-55\,\mathrm{km}$, the widening of the gap stabilises. This scale corresponds closely to the expected effective resolution of NWP models, which \textcite{skamarock_evaluating_2004} estimates at 6-10 times the grid spacing. For CERRA's 5.5 km grid, this implies an effective resolution of $\sim33-55\,\mathrm{km}$. Below this threshold, atmospheric variance is increasingly damped by the combined effects of numerical diffusion, interpolation and filtering associated with the model dynamics, physical parameterizations, and data assimilation \parencite{ridal_cerra_2024}. As a result, the reanalysis progressively loses energy at scales larger than the Nyquist wavelength \parencite{skamarock_evaluating_2004}. Because the U-Net is not subject to these stability constraints, the gap between the retrieval and CERRA narrows from this point onward. At horizontal scales of approximately $\sim17-20\,\mathrm{km}$, a crossover occurs, where the retrieval matches and often exceeds CERRA's variability. Interestingly, the crossover for 300 hPa temperature occurs at a larger scale $\sim40\,\mathrm{km}$, implying a substantially stronger increase in small-scale variability for upper-tropospheric temperature compared with the 900\,hPa retrieved temperature, or 300 and 900\,hPa retrieved humidity, an observation that invites further investigation.

This is also consistent with what was observed in the zoomed-in fields from Figures \ref{fig:retrieval_2d_300}g-j and \ref{fig:retrieval_2d_900}g-j, where the retrieval estimates fine-scale temperature and humidity structures by leveraging FCI's bands. While the reanalysis loses variance below its effective resolution due to model and assimilation damping, the retrieval expresses finer-scale variability, despite the smoothing caused by broad spectral bands, MSE optimisation, patch aggregation, and the model being trained on interpolated reanalysis. The enhanced energy below the crossover scale therefore suggests the presence of smaller scale information from the FCI observations compared to CERRA. However, in the absence of independent higher-resolution profile observations, it is not possible to determine with certainty whether this enhanced small-scale variability reflects genuine temperature and humidity structures or retrieval-induced artifacts.

\subsubsection{Spatial distribution of retrieval errors}
\begin{figure}[h!]
\centering
\includegraphics[width=1.0\linewidth]{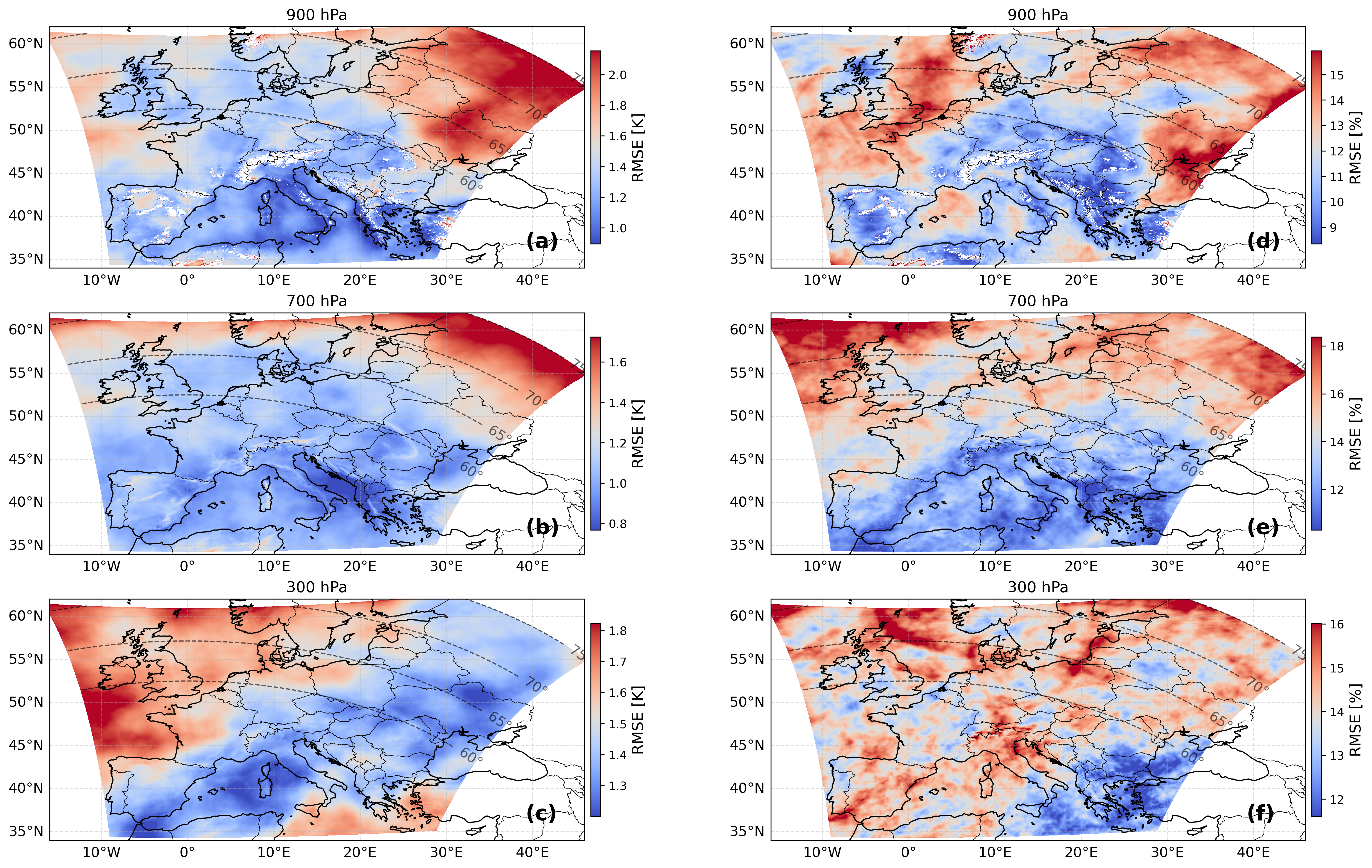}
\caption{\label{fig:rmse_map}
Test set spatial RMSE distribution of retrieved atmospheric profiles evaluated against CERRA reanalysis. 
(a-c) Temperature and
(d-f) relative humidity across three pressure levels.
Contour lines added for satellite zenith angles of 60, 65, 70 and 75°. 
}
\end{figure}

Figure \ref{fig:rmse_map} shows the spatial distribution of RMSE across the evaluation domain for temperature and relative humidity, computed against CERRA over the test period. Relative humidity is less sensitive to contrasts in absolute moisture content between coastal and continental regions, compared to specific humidity. Because CERRA serves as training, target and validation reference, systematic biases of the kind observed against radiosondes are absent, and RMSE values are lower and less variable. Spatially coherent structures that cannot be attributed to clear consistent physical mechanisms should be interpreted with caution, as with a larger evaluation set, we expect most of these structures to disappear.

For temperature (Figure \ref{fig:rmse_map}a-c), the most spatially coherent feature at 950 and 700 hPa is a progressive RMSE increase toward the northeastern corner of the domain. While regional meteorological factors may contribute to regional uncertainties, this systematic gradient is likely dominated by the increasing satellite zenith angle at higher latitudes and longitudes. At lower altitudes, the greater column of infrared-absorbing gases is geometrically amplified at oblique angles, degrading the effective signal available to the retrieval. At 300 hPa, the absorbing-gas column above is smaller and the northeastern gradient disappears. Elevated RMSE values are also visible near the Alps, Pyrenees, and Carpathians at 950 and 700 hPa, which may reflect limited abilities of FCI and CERRA to resolve temperature and humidity structures in mountain regions. 

The relative humidity RMSE distribution in Figure \ref{fig:rmse_map}d-f is less spatially coherent than temperature. A viewing-angle signature could still be suggested at 950 and 700 hPa, but less clearly than for temperature. At 300 hPa, no viewing-angle gradient is apparent in the humidity field, consistent with what is seen for temperature.

In summary, the spatial RMSE distribution points to two potential physical error sources: increasing satellite zenith angle degrading lower-tropospheric retrievals toward the northeastern domain edge, and limited spectral resolution struggling to resolve sharp gradients near complex terrain. Beyond these, the limited test set sampling makes it difficult to fully separate systematic retrieval errors from weather event mischaracterisation, a limitation that future work should address using longer evaluation periods and including comprehensive uncertainty characterisation in the retrieval.

\FloatBarrier

\subsection{Channel contributions to retrieval performance} \label{subsec:permutation_feature_sensitivity}

Different spectral regions provide sensitivity to distinct layers of the troposphere, enabling the retrieval of vertical temperature and humidity profiles from infrared radiances. This vertical sensitivity arises from wavelength-dependent atmospheric absorption, which determines the altitude at which the optical depth approaches unity and thus the effective emission level \parencite{wallace_atmospheric_2006}. Strongly absorbing channels sense emission from higher atmospheric layers, while window channels are dominated by the lower troposphere and surface \parencite{menzel_satellite-based_2018}. Unlike hyperspectral infrared sounders such as MTG-IRS, which measure on the order of thousands of narrow spectral channels, MTG-FCI is a multispectral imager with only 16 broadband channels, most of which are window channels and therefore predominantly sensitive to the Earth's surface, cloud tops, and the lower atmosphere.

\begin{figure}[h!]
\centering
\includegraphics[width=1.0\linewidth]{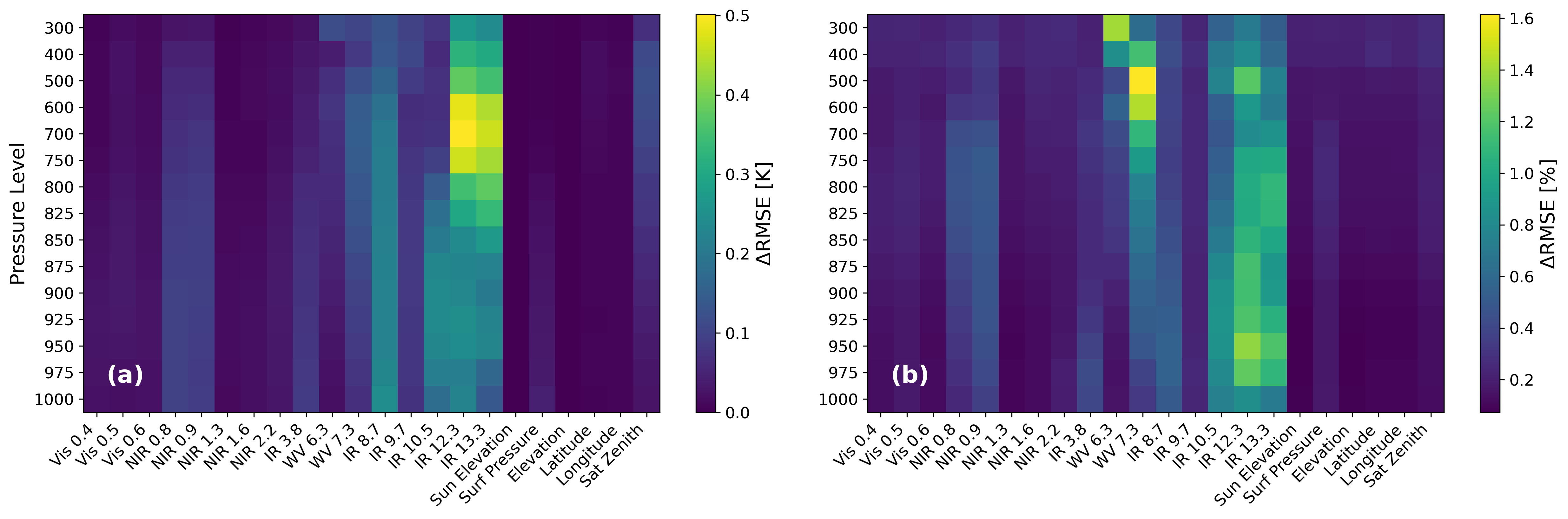}
\caption{\label{fig:feature_importance} Feature importance of the (a) temperature and (b) relative humidity profiles as the difference in RMSE compared to retrievals without substitution.}
\end{figure}

To investigate how the U-Net exploits the information contained in the FCI channels, but also the other ancillary input variables, we perform a feature importance analysis. The resulting sensitivities, shown in Figure \ref{fig:feature_importance} as RMSE increase of temperature and relative humidity, are interpreted in the context of radiative transfer theory and the vertical weighting functions of spectrally comparable channels from SEVIRI, ABI, and MODIS \parencite{schmetz_introduction_2002, seemann_operational_2003, schmit_closer_2017}.

\subsubsection{12.3\,$\mu$m and CO$_2$ channels as cloud-top anchors}

For both temperature and relative humidity retrievals, one of the most striking features is the dominant importance of the 12.3 and 13.3\,$\mu$m channels throughout the atmospheric column. Understanding this behaviour requires examining their role in cloud characterisation. In all-sky retrieval frameworks, clouds effectively become the new lower boundary condition of the radiative transfer problem \parencite{zhou_physically_2007}. Once cloud-top temperature and pressure are established, the remaining channel radiances must be interpreted relative to that new surface boundary. This is why proper characterisation of cloud top characteristics is crucial for a good retrieval.

The 13.3\,$\mu$m channel lies close to a strong CO$_2$ absorption band. Because CO$_2$ is well mixed throughout the atmosphere, radiation observed in this channel originates from higher altitudes than in window channels, resulting in substantially colder brightness temperatures \parencite{schmetz_introduction_2002}. The contrast between the 13.3\,$\mu$m channel and infrared window channels, also called radiance ratioing, therefore provides strong information on cloud-top height, particularly for semi-transparent cirrus clouds that are difficult to detect using window channels alone \parencite{hamann_remote_2014}. 

The 12.3\,$\mu$m channel provides complementary information. Unlike the nearly transparent window channel 10.5\,$\mu$m, it experiences some absorption by water vapour \parencite{schmit_closer_2017}. Consequently, the split-window difference (SWD) between 12.3 and 10.5\,$\mu$m is sensitive to both cloud properties and atmospheric water vapour, making it valuable for distinguishing humidity variations from cloud-induced radiance changes. Furthermore, much like the 13.3\,$\mu$m band, the 12.3\,$\mu$m channel is also directly utilized in radiance ratioing to mathematically decouple cloud-top height from cloud emissivity for clouds in the lower-troposphere \parencite{hamann_remote_2014}.

Accurately anchoring the new surface boundary condition across all cloud types requires a combination of channels: while a clean window band is sufficient to estimate the top of opaque overcast clouds via fitting the cloud top brightness temperature to forecast profiles, the 13.3\,$\mu$m and 12.3\,$\mu$m bands are essential to correct moisture-affected brightness temperature and for semi-transparent and partially clouded pixels \parencite{hamann_remote_2014}. Together, they provide the necessary information to correctly anchor the atmospheric column under any cloudy condition. In the feature importance analysis, shuffling either channel introduces an internally inconsistent scene: the perturbed channel implies a different cloud-top state than the remaining observations. Errors in cloud-top placement or temperature then propagate throughout the retrieved profile. For the temperature profiles, the impact is particularly large in the upper troposphere, probably because FCI lacks dedicated temperature-sounding channels and is more sensitive to cloud characteristics and lower boundary conditions. Additionally, while the two water-vapour bands (7.3 and 6.3\,$\mu$m) provide direct sensitivity to mid-upper tropospheric moisture, the 12.3\,$\mu$m channel helps constrain how water vapour is distributed vertically relative to the total column and cloud layer. This could explain the similarity between the vertical relative humidity RMSE increase associated with perturbing the 12.3 and 7.3\,$\mu$m channels between 600 to 400\,hPa. Overall, because our substitution approach shuffles patches across varying cloud conditions, corrupting these two bands seems to introduce a scene-wide inconsistency that affects the entire atmospheric column of both retrieved profiles.

\subsubsection{Other infrared channels}
Beyond the cloud-top anchor bands, the thermal infrared channels exhibit sensitivities broadly consistent with radiative transfer principles.
The 10.5\,$\mu$m clean window channel shows substantial contribution across the full profile for both variables. Its primary role is as the brightness-temperature baseline against which moisture-affected channels are differenced: a larger contrast between the 10.5\,$\mu$m brightness temperature and an adjacent absorbing channel indicates greater moisture in the sensitive layer. It also provides direct constraints on cloud-top and sea-surface temperature and on snow cover extent. While the clean-window band is important in estimating the cloud top temperature, the other window bands could compensate for the loss of information when shuffled, which explains its lower importance compared to the 12.3 and 13.3\,$\mu$m bands.
The 9.7\,$\mu$m ozone band shows small contribution concentrated in lower and upper tropospheric temperature. Its SEVIRI-equivalent weighting function peaks near 100\,hPa, above the nominal retrieval domain, but tropopause dynamics and ozone-induced brightness temperature cooling could account for the residual upper-level sensitivity observed here.
The 8.7\,$\mu$m surface-sensitive window channel, like the 10.5\,$\mu$m window, shows substantial contribution to mid-to-lower tropospheric temperature and moisture, consistent with MODIS Band\,29 weighting function, which shows clear lower-tropospheric water vapour sensitivity \parencite{seemann_operational_2003}.
The 7.3 and 6.3\,$\mu$m water vapour bands exhibit the largest humidity contribution, and physically consistent sensitivities in the humidity field: the 7.3\,$\mu$m channel peaks in the mid-to-upper troposphere and the 6.3\,$\mu$m channel in the upper troposphere, in close agreement with SEVIRI weighting functions \parencite{schmetz_introduction_2002}. The 6.3\,$\mu$m weighting function extends to approximately 200\,hPa, suggesting that the retrieval domain could be extended to higher altitudes in future work.
The 3.8\,$\mu$m shortwave band occupies a unique position, sensing both emitted terrestrial and reflected solar radiation. This mixed signal is notoriously difficult to simulate in fast radiative transfer models owing to the bidirectional solar path, variable surface reflectance, and coupled scattering effects \parencite{gao_water_2003}, and for this reason is for example excluded from the operational ABI profile retrieval \parencite{schmit_legacy_2019}. Despite this, the U-Net extracts substantial sensitivity from this channel for lower-tropospheric temperature and moisture, a clear illustration of the data-driven approach's capacity to exploit information that is difficult to incorporate in physical inversion schemes.

\subsubsection{Visible and near-infrared channels}
The near-infrared bands at 2.2, 1.6, and 1.3 $\mu$m show relatively small contributions. The 2.2 and 1.6 $\mu$m bands primarily characterise cloud phase, particle size, and aerosols, with limited direct thermodynamic sensitivity. The 1.3 $\mu$m band has extremely strong water vapour absorption, so strong that the signal rarely reaches the surface, and consequently shows the smallest contribution to moisture retrieval of all FCI channels \parencite{schmit_closer_2017}.
The 0.9 and 0.8 $\mu$m bands show the strongest contributions among all reflected channels, and their joint sensitivity seems to tell a physically coherent story. The 0.9 $\mu$m band retrieves near-surface water vapour by measuring the attenuation of solar radiation along its round-trip path through the atmosphere \parencite{gao_water_2003}. Because most atmospheric moisture resides in the lowest layers, this bidirectional path gives the channel strong sensitivity to near-surface humidity. However, a single absorption channel cannot distinguish between high atmospheric moisture and a naturally dark surface. The 0.8 $\mu$m vegetation band resolves this ambiguity. As an atmospheric window reference, it carries small gaseous absorption signal, and using the ratio between the two channels cancels the highly variable effects of surface reflectance to isolate the moisture signal \parencite{el_kassar_optimal_2026}. This is most often used for lower-tropospheric retrieval in clear-sky, but the same principle works above clouds. The comparable and jointly elevated sensitivity of both bands suggests that our retrieval successfully exploits this physical relationship, a result that is particularly encouraging given that this ratio-based moisture retrieval is not explicitly encoded in the training procedure.
The visible bands at 0.6, 0.5, and 0.4 $\mu$m show very small contributions. Without gaseous absorption, their direct thermodynamic sensitivity is limited, but they could still theoretically be providing some contextual information on cloud boundaries, land cover, and aerosols that helps anchor retrievals near the surface.

\subsubsection{Further predictors}
Sun elevation, surface elevation, latitude, and longitude all show negligible sensitivity. Sun elevation may be estimated from visible and near-infrared reflectances; latitude and longitude are correlated with satellite zenith angle, surface elevation; and terrain elevation is largely encoded in surface pressure. Shuffling these quantities individually has little effect on performance, potentially because the model may be able to infer these quantities from each other. Surface pressure, the one ancillary variable derived from reanalysis, shows a small but consistent contribution to lower-to-mid tropospheric temperature and humidity. The satellite zenith angle shows the most relevant ancillary contribution, with sensitivity spanning the full profile but peaking at upper levels. This seems consistent with infrared radiative transfer Jacobians shifting upward as the satellite zenith angle increases \parencite{eumetsat_mtg-irs_2021}, and indirectly with the RMSE spatial distribution (Figure \ref{fig:rmse_map}), showing that higher satellite zenith angles degrade lower tropospheric retrievals. The higher altitude sensitivity seems to suggest model awareness of the influence of the satellite zenith angle on atmospheric path length and parallax displacement. 

The feature importance analysis suggests that the retrieval is broadly physically consistent with established radiative transfer characteristics of FCI bands. The important role of the 12.3 and 13.3 $\mu$m bands likely reflects the central importance of cloud-top characterisation as the lower boundary condition in all-sky conditions, amplified by the scene inconsistencies introduced through cross-sample substitution. The strong sensitivity to the 0.8/0.9 $\mu$m pair and the 3.8 $\mu$m band illustrates the ability of the data-driven approach to exploit physically grounded relationships that are less commonly incorporated in traditional retrieval schemes. Finally, the relatively small contribution of several ancillary variables suggests that the U-Net can implicitly derive contextual information that would otherwise require explicit inputs.

\subsection{Comparison to other retrievals}

To contextualise our results, Table \ref{tab:retrievals_comparison} compares our FCI retrieval framework against reported performances of two established operational products: the Infrared Atmospheric Sounding Interferometer (IASI) aboard the polar-orbiting Metop series, and the Advanced Baseline Imager (ABI) aboard the geostationary GOES-16 satellite. While instruments and evaluation methodologies differ substantially, this order-of-magnitude context shows our retrieval achieves error margins comparable to both instruments, despite important methodological differences that should be kept in mind when interpreting these comparisons. 

The datasets differ significantly in their spectral, spatial, and temporal characteristics, which impacts the strictness of their collocation criteria. First, IASI benefits from 8,461 spectral channels, providing substantially more information on the atmospheric state than the 16 channels available to our FCI-based retrieval. Furthermore, the IASI validation features a coarser spatial resolution ($\sim12\,\mathrm{km}$) and adopts different collocation criteria, requiring that the center of the observed pixel falls within 50\,$\mathrm{km}$ and 3 hours of the radiosonde measurements. In contrast, our retrieval enforces collocation within $\sim5\,\mathrm{km}$ of the launch site and within $\sim45\,minutes$ of the radiosonde measurements (Figure S4). The difference in collocation could slightly inflate errors from IASI and underestimate its performance compared to our retrieval. However, the IASI all-sky evaluation also draws on a combined infrared and microwave sounder retrieval, providing physically measured constraint in cloudy conditions, not available in our retrieval. For the geostationary comparison, the ABI retrieval \parencite{schmit_legacy_2019} shares our spatial resolution (2\,$\mathrm{km}$ at nadir) and employs closer collocation criteria, using a $\le$ 0.2$^\circ$ radius and a 30-minute window relative to the radiosonde measurements. However, the underlying retrieval methodologies differ fundamentally. The ABI retrieval relies on short-range numerical weather prediction (NWP) forecasts from the Global Forecast System (GFS) as a background prior to guide its physical inversion, and it is strictly constrained to clear-sky conditions. Our retrieval, by contrast, operates in all-sky conditions without using any prior NWP information, meaning its performance faces the natural penalties associated with cloud attenuation and a lack of a guided background state.

\begin{table}[ht]
\centering
\begin{tabular}{llll}
\hline
Variable & IASI & ABI & FCI U-Net \\
\hline
Temperature bias & -1 to 0.2\,K & -0.4 to 0.2\,K & -0.3 to 0.1\,K \\
Temperature STD & 1.0 to 2.0\,K & 0.7 to 2.0\,K & 1.5 to 1.9\,K \\
Specific humidity bias & -0.1 to 0.2\,g/kg & - & -0.03 to 0.13\,g/kg \\
Specific humidity STD & 0.0 to 1.2\,g/kg & - & 0.04 to 1.21\,g/kg \\
Relative humidity bias & - & 0 to 16\,\% & 1 to 25\,\% \\
Relative humidity STD & - & 11 to 15\,\% & 12 to 20\,\% \\
\hline
\end{tabular}
\caption{Comparison of our FCI retrievals with IASI and ABI bias and STDs against radiosondes. IASI and ABI bias and STDs were approximated: for IASI from Figures 30-31 of the IASI level 2 validation report \parencite{eumetsat_validation_2021}, and for ABI from \textcite{schmit_legacy_2019} validation report for GOES-16 Figure 2.}
\label{tab:retrievals_comparison}
\end{table}

Despite these structural and data constraints, our FCI retrievals demonstrate a statistical reliability that is somewhat comparable to both IASI and ABI. Crucially, they do so using only the 16 broadband channels of MTG-FCI, without the thousands of spectral bands available to IASI and without injecting NWP forecast data to stabilise the solution. The higher relative humidity biases relative to ABI are consistent with the reanalysis-inherited moist bias structure discussed in Section 4.1.1. One future work direction could be to investigate whether incorporating NWP forecast information as an additional input would further constrain the retrieval and improve profile accuracy, without relying too much on the background forecast.

\section{Conclusions}
This study demonstrates that a fully data-driven, spatially aware deep learning framework can retrieve statistically reliable all-sky tropospheric temperature and humidity profiles from MTG-FCI observations without relying on NWP forecast backgrounds. Trained on 14 months of collocated FCI observations and CERRA reanalysis profiles, our U-Net-based retrieval achieves temperature biases and standard deviations of -0.3 to 0.1\,K and 1.5 to 1.9\,K, and relative humidity biases and standard deviations of 1 to 25\,\% and 12 to 20\,\% against independent radiosonde observations across most of the troposphere, outperforming the 30-year ERA5 climatology, the ancillary-only model and the pixel-wise 1x1-Net, while substantially closing the performance gap to CERRA reanalysis.

The retrieval performance varies predictably with observational conditions. Temperature STDs increase modestly below cloud tops and during nighttime, but the retrieval maintains consistent performance. The retrieval can draw on above-cloud retrieved temperature and humidity structure, spatial context — as demonstrated by the pixel-wise comparison — and possibly partial radiative transmission through optically thin clouds to estimate below cloud tops. Daytime retrievals show positive contribution from visible and near-infrared channels, which have sensitivity to lower-tropospheric moisture \parencite{gao_water_2003, zhou_physically_2007, schmit_legacy_2019}. Seasonal variability is modest and consistent with patterns reported for the IASI infrared sounder over Europe, with winter retrievals showing higher STDs driven by stronger baroclinic activity, increased cloud cover, and lower sun elevations. Despite the limited FCI record, an evaluation on an independent three-months period outside the training window confirms that the model generalises well through a large part of the troposphere, with some degradation in the near-surface boundary layer and at 300–400\,hPa.

The spatial and power spectra analyses confirm that large-scale temperature and humidity structures are well reproduced. Below CERRA's effective resolution, the retrieval recovers fine-scale variance from FCI's bands with 2\,$\mathrm{km}$ resolution at nadir that the reanalysis cannot represent. At horizontal scales of 17 to 550\,$\mathrm{km}$, however, gradients are systematically smoothed, which we see as a consequence of the limited imager spectral resolution and the MSE optimisation, which penalises sharp spatially offset features and produces attenuated averages at the expense of sharpness. Retrieval errors increase toward higher satellite zenith angles, where oblique viewing geometry amplifies atmospheric path lengths, and over high-elevation areas..

The feature importance analysis shows sensitivity patterns broadly consistent with known radiative transfer characteristics of the FCI bands, despite being learned entirely from data without explicit physical constraints. The 12.3 and 13.3 µm channels dominate throughout the column, reflecting the central role of cloud-top characterisation as the effective lower boundary condition in all-sky retrievals, while the contributions of the 0.9/0.8 µm pair and the 3.8 µm band illustrate the approach's capacity to exploit physically grounded information rarely used in operational inversion schemes.

Our results establish that modern deep learning architectures, combined with the spatial and spectral information of MTG-FCI, can provide fast, forecast-independent temperature and humidity profiles at high spatial and temporal resolution. This opens a path toward more frequent and autonomous tropospheric monitoring with the potential to complement sounder-based retrievals. 

\section{CRediT authorship contribution statement}

\textbf{Alejandro Salgueiro}: Writing - original draft, Visualization, Validation, Methodology, Investigation, Formal analysis, Data curation, Conceptualization. \textbf{Angela Meyer}: Writing - review and editing, Supervision, Resources, Methodology, Funding acquisition, Conceptualization. \textbf{Johannes Rausch}: Writing - review and editing, Methodology, Conceptualization. \textbf{Julie Thérèse Villinger}: Writing - review and editing, Methodology, Conceptualization.

\section{Declaration of competing interest}
The authors declare the following financial interests/personal relationships which may be considered as potential competing interests: J.V. and J.R. were employed by the company Meteomatics during this study. The other authors declare that they have no known competing financial interests or personal relationships that could have appeared to influence the work reported in this paper.

\section{Acknowledgements}
This work was funded by the Swiss Innovation Agency Innosuisse.

\paragraph{AI tools}
Generative AI (Claude Sonnet 4.6, Anthropic) was used during the preparation of this manuscript to improve language clarity and text readability.

\section{Open research}
\paragraph{Availability statement}
The Flexible Combined Imager (MTG-FCI) Level 1c observations are available in EUMETSAT Data Store at \url{https://data.eumetsat.int/product/EO:EUM:DAT:0662} \parencite{eumetsat_fci_2024}, while FCI's Optimal Cloud Analysis is available at \url{https://data.eumetsat.int/product/EO:EUM:DAT:0684} \parencite{eumetsat_optimal_2025}. CERRA reanalysis profiles are available in Copernicus Climate Data Store at \url{https://cds.climate.copernicus.eu/datasets/reanalysis-cerra-pressure-levels} \parencite{copernicus_climate_change_service_cerra_2022}, while ERA5 monthly averaged climatology is available at \url{https://cds.climate.copernicus.eu/datasets/reanalysis-era5-pressure-levels-monthly-means} \parencite{copernicus_climate_change_service_era5_2019}. We also used ERA5 Land and ERA5 surface pressure for the second test set, when CERRA was not available. These are available at \url{https://cds.climate.copernicus.eu/datasets/reanalysis-era5-land} \parencite{copernicus_climate_change_service_era5-land_2019} and \url{https://cds.climate.copernicus.eu/datasets/reanalysis-era5-single-levels} \parencite{copernicus_climate_change_service_era5_2018} respectively. The radiosonde measurements are available in NOAA Integrated Global Radiosonde Archive (IGRA) at \url{https://www.ncei.noaa.gov/products/weather-balloon/integrated-global-radiosonde-archive} \parencite{durre_integrated_2016}.

\printbibliography

\end{document}